\documentclass{article}

\PassOptionsToPackage{round}{natbib}

\usepackage[main, final]{neurips_2026}

\usepackage[utf8]{inputenc} % allow utf-8 input
\usepackage[T1]{fontenc}    % use 8-bit T1 fonts
\usepackage{hyperref}       % hyperlinks
\usepackage{url}            % simple URL typesetting
\usepackage{amsmath}        % advanced math environments and symbols
\usepackage{booktabs}       % professional-quality tables
\usepackage{amsfonts}       % blackboard math symbols
\usepackage{nicefrac}       % compact symbols for 1/2, etc.
\usepackage{microtype}      % microtypography
\usepackage[table]{xcolor}  % colors and table cell highlighting
\usepackage{multirow}
\usepackage{array}
\usepackage{graphicx} % Required for resizebox
\usepackage{longtable}
\usepackage{caption}
\usepackage{subcaption}
\usepackage{wrapfig}
\usepackage{afterpage}
\usepackage{etoc}
\AtBeginDocument{\etocdepthtag.toc{main}}
\usepackage{fvextra}
\DefineVerbatimEnvironment{promptbox}{Verbatim}{breaklines=true,breakanywhere=true,fontsize=\footnotesize,bgcolor=promptgray,framesep=2mm}

\title{Need-Aware Multi-Objective Reinforcement Learning for Emotionally Intelligent LLM Agents}

\author{%
  Xiaohe Bo$^{1*}$\setcounter{footnote}{2}\thanks{Work done during an internship at Huawei.\newline\makebox[1.8em][r]{$^{*}$}Equal contribution.\quad $^{\dagger}$Co-corresponding authors.}, Xueyang Feng$^{1}$\footnotemark[1], Rui Li$^{1}$, Zihang Tian$^{1}$, Xu Chen$^{1 \dagger}$, Quanyu Dai$^{2 \dagger}$\\
  $^{1}$Gaoling School of Artificial Intelligence, Renmin University of China\\
  $^{2}$Huawei Technologies Ltd.\\
  \texttt{\{xiaohe,xu.chen\}@ruc.edu.cn}\quad \texttt{daiquanyu@huawei.com}
}

\begin{document}

% Distinct highlight colors for the top three ranks.

\definecolor{best}{HTML}{C4DBF1}       % Best
\definecolor{secondbest}{HTML}{DCEAF7} % Second best
\definecolor{thirdbest}{HTML}{F0F6FC}  % Third best
\definecolor{gbest}{HTML}{DDD3F2}
\definecolor{gsecondbest}{HTML}{EDE7F8}
\definecolor{gthirdbest}{HTML}{F7F3FC}

\definecolor{best1}{HTML}{ffced2}       % Best
\definecolor{secondbest1}{HTML}{fedee1} % Second best
\definecolor{thirdbest1}{HTML}{ffebed}  % Third best
\definecolor{casestatepink}{HTML}{FFF5F6}
\definecolor{mosteffective}{HTML}{F6DDE8}
\definecolor{secondeffective}{HTML}{FBEFF5}
\definecolor{promptgray}{HTML}{FAFAFA}

\maketitle
\vspace{-16pt}

\begin{abstract}
In emotional support conversations, user needs are dynamic, shifting across turns between emotional soothing, cognitive guidance, and their combinations. Existing methods often optimize a single user-state objective, such as emotional relief or cognitive growth, making it difficult to capture such need shifts. In this paper, we formulate emotional support dialogue as a need-aware multi-objective optimization problem, where agents learn to generate responses that balance multiple support objectives according to users' evolving needs. We propose \textbf{MindFlow}, a psychologically grounded user simulator that tracks users' emotional and cognitive needs at each turn, derives state-dependent preference vectors from personalized arousal curves, and models mental-state transitions with a dual-process system. Building on MindFlow, we introduce \textbf{NeMo}, a need-aware multi-objective reinforcement learning framework that estimates local multi-turn effects through forward simulation and dynamically aggregates emotional and cognitive rewards based on current support preferences. By converting delayed, multi-dimensional feedback into step-level optimization signals, NeMo improves credit assignment and aligns policy learning with long-term user-state improvement. Experiments show that NeMo consistently improves user-side state metrics and response-level supportive quality, with robust generalization under both Sim2Sim and Sim2Real evaluations. 
\end{abstract}

\section{Introduction}

As large language models (LLMs) continue to improve in emotional reasoning~\citep{ref:emotion_reasoning1,ref:emotion_reasoning2,ref:emotion_reasoning3,ref:emotion_reasoning4} and social understanding~\citep{ref:social_reasoning1,ref:social_reasoning2,ref:social_reasoning3,ref:social_reasoning4,ref:social_reasoning5}, their applications are expanding from task-oriented assistance to human-centered scenarios such as emotional support conversations (ESC). In ESC, an effective agent must go beyond factual correctness or surface-level empathy: it should generate responses that are aligned with the user's emotional readiness, cognitive state, and evolving support needs across a multi-turn interaction~\citep{ref:esconv,ref:sage,ref:rlver,ref:socialsim}. This ability is often regarded as a form of emotional intelligence.

Early efforts to improve the emotional intelligence of LLMs mainly relied on human-collected dialogues~\citep{ref:esconv,ref:psyqa,ref:xinling} or static data synthesis~\citep{ref:aug_esc,ref:chatcounselor,ref:extes}, optimizing text-side supportive skills such as empathy, question, and suggestion. More recent work has shifted toward user-centered optimization, where user simulators are constructed to infer psychological state changes during multi-turn dialogues and provide reinforcement learning signals~\citep{ref:dpdp}. For example, RLVER~\citep{ref:rlver} and PPDPP~\citep{ref:ppdpp} optimize emotion-related feedback, while ES-VR~\citep{ref:esvr} focuses on cognitive growth. Compared with text-side supervision, these methods more directly target user-state improvement and therefore better align with the goal of emotional support.

\begin{figure}[!t]
    \centering
    \setlength{\abovecaptionskip}{0.03cm}
    \includegraphics[width=0.92\linewidth]{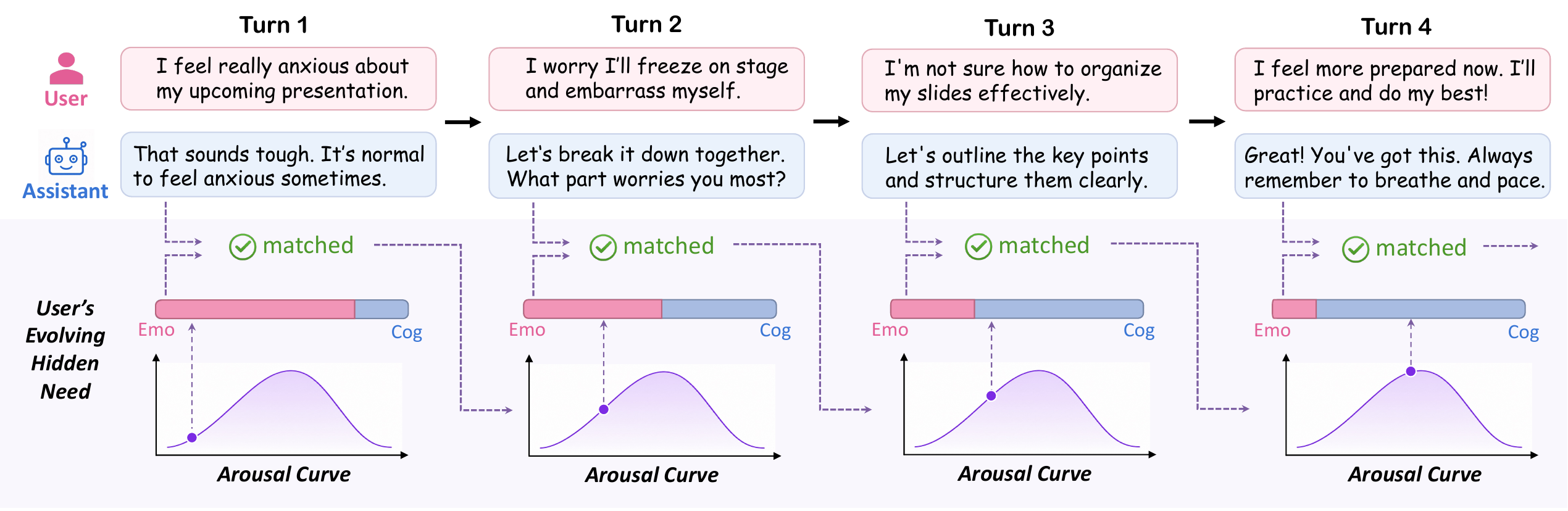}
    \caption{
    Effective emotional support should match users' evolving needs across dialogue turns.
    }
    \label{fig:intro}
    \vspace{-0.3cm}
\end{figure}

However, existing user-centered methods often reduce emotional intelligence to a single user-state objective, such as emotional relief or cognitive growth. In practice, user needs are dynamic and multi-dimensional: a user may first need emotional soothing to reduce distress, then become ready for cognitive guidance such as reframing or action planning~\citep{ref:esconv,ref:arousal_theory}. Premature cognitive guidance may trigger defensiveness when the user is still emotionally unsettled, while purely affective comfort may become insufficient once the user is prepared to reflect and act~\citep{ref:cbt,ref:matching_theory}. As illustrated in Figure~\ref{fig:intro}, effective emotional support requires agents to dynamically balance multiple support objectives and match users' evolving needs across turns.

Motivated by this observation, we formulate emotional support dialogue as a need-aware multi-objective decision-making problem~\citep{ref:multi_objective_rl,ref:rl_introduction}. Rather than optimizing a fixed reward, the agent should balance the emotional relief and cognitive growth objectives according to the user's current support preference. This formulation raises two key challenges. First, a simulator should explicitly model the coexistence of emotional and cognitive needs and constrain the evolution of their relative priorities with psychologically plausible dynamics, rather than treating these priorities as arbitrary hidden variables. Second, policy learning must convert delayed, multi-dimensional user feedback into reliable step-level optimization signals. A response may affect emotional relief and cognitive growth over several subsequent turns, and these effects can evolve asynchronously. How to design step-level rewards and reliably guide policy learning remains an open question.

To address these challenges, we propose \textbf{MindFlow}, a psychologically grounded user simulator, and \textbf{NeMo}, a need-aware multi-objective reinforcement learning framework for emotionally intelligent LLM agents. 
MindFlow explicitly tracks users' emotional and cognitive states, derives support preferences from personalized arousal curves inspired by optimal arousal theory~\citep{ref:yerkes_dodson,ref:arousal_hebb,ref:arousal_theory}, and simulates mental-state transitions with a dual-process mechanism, thereby providing multi-dimensional user-side feedback. 
Building on MindFlow, NeMo derives need-aware step-level rewards with two modules: forward simulation estimates each response's local multi-turn effect, and dynamic aggregation combines emotional and cognitive rewards according to the current support preference. These rewards are used in online GRPO training~\citep{ref:grpo}, guiding the policy to balance emotional and cognitive objectives according to user needs.

Our contributions are summarized as follows:
(1) We formulate emotional support dialogue as a need-aware multi-objective optimization problem, where agents must dynamically balance emotional relief and cognitive growth objectives across turns as users' needs evolve.
(2) We introduce \textbf{MindFlow}, a psychologically grounded simulator that models emotional and cognitive states, derives preference vectors from personalized arousal curves, and updates mental states through a dual-process mechanism.
(3) We propose \textbf{NeMo}, a reinforcement learning framework that converts MindFlow feedback into need-aware step-level rewards via multi-turn forward simulation and dynamic reward aggregation.
(4) Experiments show that NeMo improves both user-state metrics and response-level supportive quality, with robust generalization under Sim2Sim and Sim2Real evaluations.

\section{Problem Formulation}
\label{sec:problem_formulation}
Following the need-aware motivation above, we formalize emotionally intelligent dialogue as a multi-objective sequential decision-making problem driven by the user's latent mental state. At turn $t$, the user produces an utterance $u_t$, and the system observes the dialogue history $s_t=(u_0,a_0,\ldots,u_{t-1},a_{t-1},u_t)$ and generates a response $a_t\sim\pi_\theta(\cdot\mid s_t)$. Unlike surface-level dialogue modeling, emotional dialogues require accurately modeling the user's latent mental state $M_t$. The user utterance $u_t \sim p(\cdot \mid M_t, s_{t-1}, a_{t-1})$ can thus be viewed as an observable manifestation of $M_t$ under current context. Therefore, the trajectory of a $T$-turn dialogue can be written as:
\begin{equation}
\xi=(M_0,u_0,a_0,M_1,u_1,a_1,\ldots,M_T,u_T,a_T,M_{T+1}).
\end{equation}
Based on the state transition, the simulator provides feedback after each turn. Instead of modeling state improvement from a single dimension, we define the turn-level feedback as a multi-objective reward $r_t=(r_t^{emo},r_t^{cog})$, with a preference vector $w_t=(w_t^{emo},w_t^{cog})$ to reflect the relative importance of each objective. Since the initial reward $r_0$ is fixed by the user profile and scenario, maximizing user-state improvement is equivalent to maximizing the final-turn reward, which can be written as:
\begin{equation}
J(\pi_\theta)
=
\mathbb{E}_{\xi\sim \pi_\theta}
\left[
\left(
r_T^{emo},
r_T^{cog}
\right)
\right].
\end{equation}

\section{MindFlow: Dynamic Multi-Objective User Simulation}
\begin{figure}[!t]
    \centering
    \setlength{\abovecaptionskip}{0.15cm}
    \includegraphics[width=0.95\linewidth]{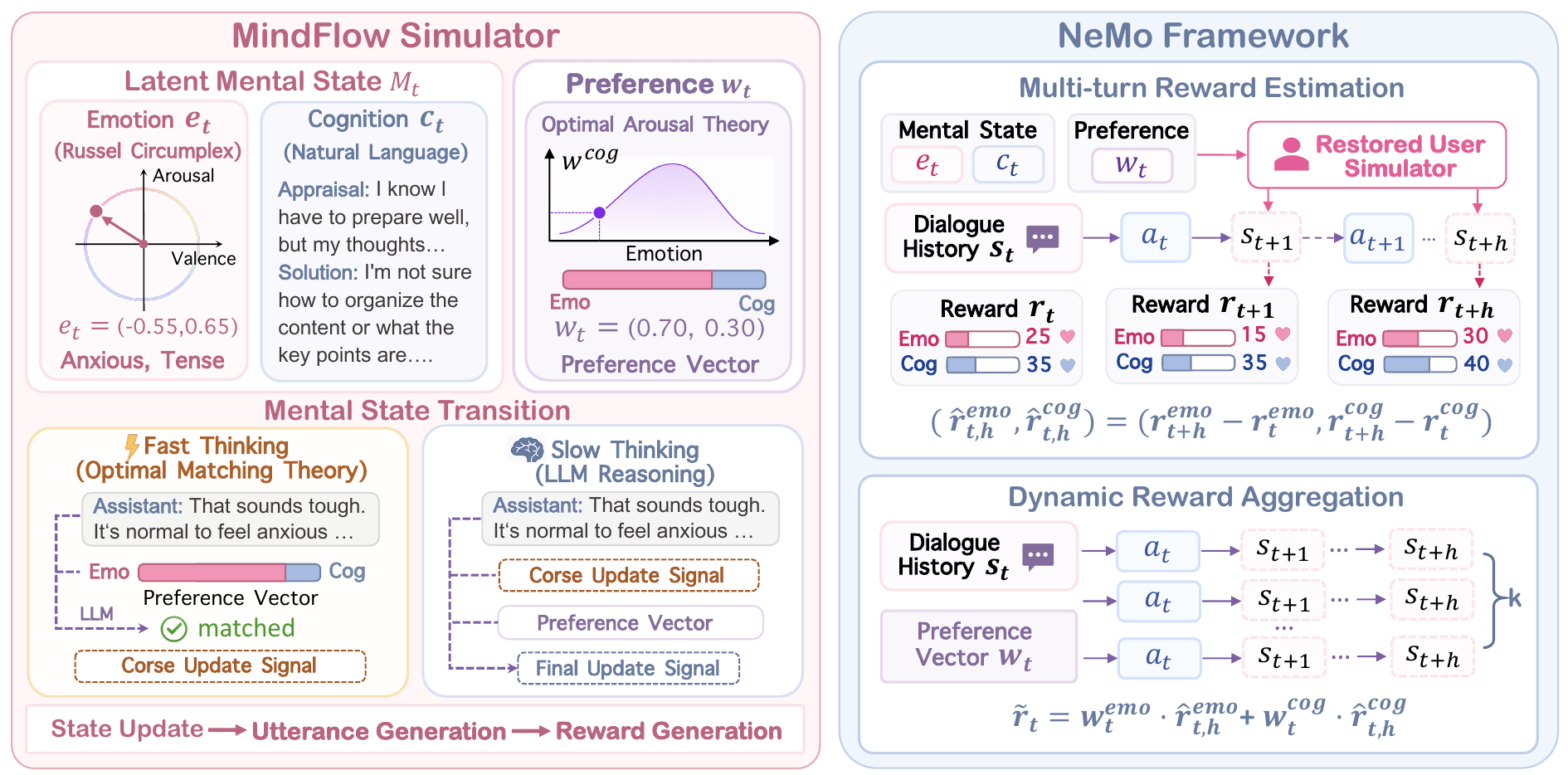}
    \caption{
    Overview of our proposed MindFlow simulator and NeMo framework. MindFlow simulates users' evolving emotional and cognitive states and derives turn-level support preferences, while NeMo uses multi-turn reward estimation and dynamic reward aggregation to optimize responses according to current user needs.
    }
    \label{fig:method_framework}
    \vspace{-0.3cm}
\end{figure}

In this section, we introduce \textbf{MindFlow}, a psychologically grounded user simulator that models evolving emotional and cognitive support needs during dialogue. Following the hidden-trajectory formulation in Section~\ref{sec:problem_formulation}, we decompose the simulation into two coupled processes: (1) $M_t \rightarrow w_t$, deriving a state-dependent support preference vector from the current mental state through personalized arousal curves (Section~\ref{sec:need_prioritization}); and (2) $M_t \rightarrow M_{t+1}$, modeling mental-state transitions after a system response through a dual-process mechanism to generate human-aligned user utterances and multi-dimensional rewards (Section~\ref{sec:mental_evolution}). We provide an overview of MindFlow in Figure~\ref{fig:method_framework} and denote all simulator components collectively by $\phi$ in the following sections for convenience.

\subsection{Arousal-Constrained Support Preference Modeling}
\label{sec:need_prioritization}

\textbf{Mental State Representation.}
To characterize emotional and cognitive support objectives, we define the user's latent mental state as $M_t=(e_t,c_t)$, where $e_t$ and $c_t$ denote the emotional and cognitive states, respectively. Following prior work~\citep{ref:emotion_representation,ref:emotion_representation2}, we represent $e_t$ as a continuous two-dimensional vector based on the Russell circumplex model~\citep{ref:russell_circumplex}, and describe $c_t$ in natural language. A concrete example of $M_t$ is shown in the upper-left part of Figure~\ref{fig:method_framework}. From $M_t$, MindFlow infers a support preference vector $w_t=(w_t^{emo},w_t^{cog})$, which specifies the relative support preference between emotional soothing and cognitive guidance.

\textbf{Support Preference Modeling.}
Optimal arousal theory~\citep{ref:yerkes_dodson,ref:arousal_theory,ref:arousal_hebb} suggests an inverted-U relationship between a user's emotional arousal and cognitive receptivity. As illustrated in Figure~\ref{fig:method_framework}, excessively low or high arousal reduces receptivity to cognitive support, whereas moderate arousal makes users more open to cognitive guidance. MindFlow thus derives the turn-level support preference vector $w_t$ from a personalized arousal curve $f_u$.

Specifically, we parameterize the personalized arousal curve $f_u$ with a Beta distribution, which provides a flexible inverted-U shape over normalized arousal. We then feed the normalized arousal value $\bar e_t$ of the emotional state $e_t$ into $f_u$ to obtain the cognitive-support weight, and define the emotional-support weight as its complement:
\begin{equation}
w_t^{cog}=
\frac{\mathrm{Beta}(\bar e_t;\alpha_u,\beta_u)}
{\max_{\bar e\in[0,1]}\mathrm{Beta}(\bar e;\alpha_u,\beta_u)},
\qquad
w_t^{emo}=1-w_t^{cog}.
\end{equation}
Here, $(\alpha_u,\beta_u)$ are user-specific parameters controlling the personalized arousal curve. Details of state definition and curve construction are provided in Appendix~\ref{sec:state_definitions} and Appendix~\ref{sec:mindflow_construction}.

\subsection{Dual-Process Mental-State Transition}
\label{sec:mental_evolution}

To model the user's mental-state transition, we build a fast--slow thinking system inspired by dual-process theory~\citep{ref:thinking_fast_and_slow,ref:dual_processing}. The fast system captures coarse response-preference alignment as an immediate affective reaction, and the slow system refines it into final emotional and cognitive state changes for generating user utterances and multi-dimensional rewards.

\textbf{Fast Thinking System.}
The fast system is built on optimal matching theory~\citep{ref:matching_theory} to measure the alignment between the agent response and user support preference. Given the agent response $a_t$, and support preference vector $w_t$, the system outputs an initial update signal:
\begin{equation}
\tilde{\Delta}_t=(\tilde{\Delta}_t^{emo},\tilde{\Delta}_t^{cog})
=
T_{\mathrm{fast}}^{\phi}(a_t,w_t).
\end{equation}
\textbf{Slow Thinking System.}
The slow-thinking system then refines the coarse signal $\tilde{\Delta}_t$ with richer dialogue context and delicate reasoning, while preserving the preference-response constraint from the fast system. This produces the final emotional--cognitive update signal:
\begin{equation}
\Delta_t=(\Delta_t^{emo},\Delta_t^{cog})
=
T_{\mathrm{slow}}^{\phi}(s_t,a_t,w_t,\tilde{\Delta}_t).
\end{equation}
The latent state is then updated as $M_{t+1}=F_{\phi}(M_t, a_t, \Delta_t)$, after which $u_{t+1}$ and $r_t$ are generated:
\begin{equation}
u_{t+1} = P_{\phi}(M_{t+1},s_t,a_t),
\qquad
r_t=R_{\phi}(M_t,M_{t+1}).
\end{equation}

\subsection{Reliability Verification}
\begin{figure*}[!t]
    \centering
    \setlength{\abovecaptionskip}{0.15cm}
    \begin{minipage}[t]{0.24\textwidth}
        \centering
        \includegraphics[width=\linewidth]{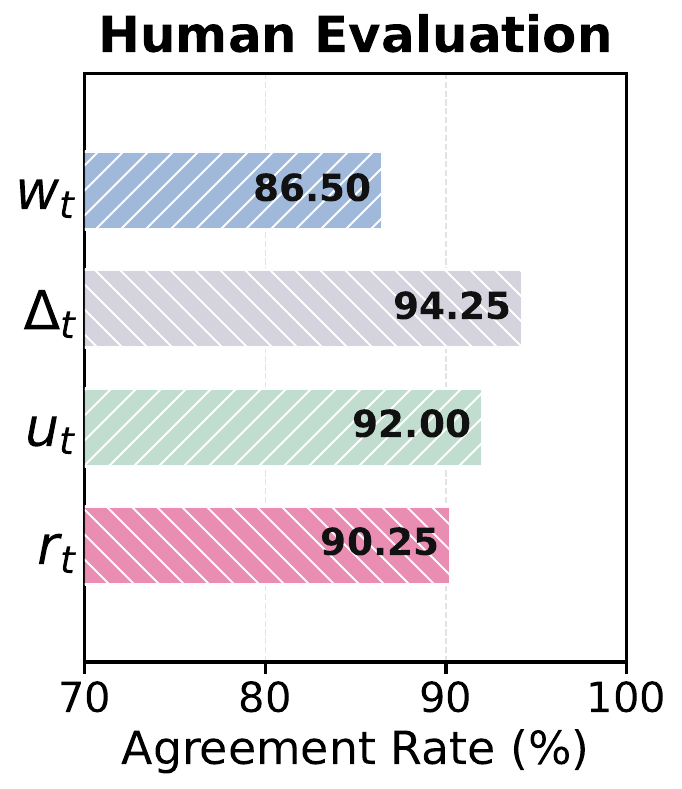}
        \caption{
        Human validation results of MindFlow.
        }
        \label{fig:eval_results}
    \end{minipage}
    \hfill
    \begin{minipage}[t]{0.72\textwidth}
        \centering
        \includegraphics[width=\linewidth]{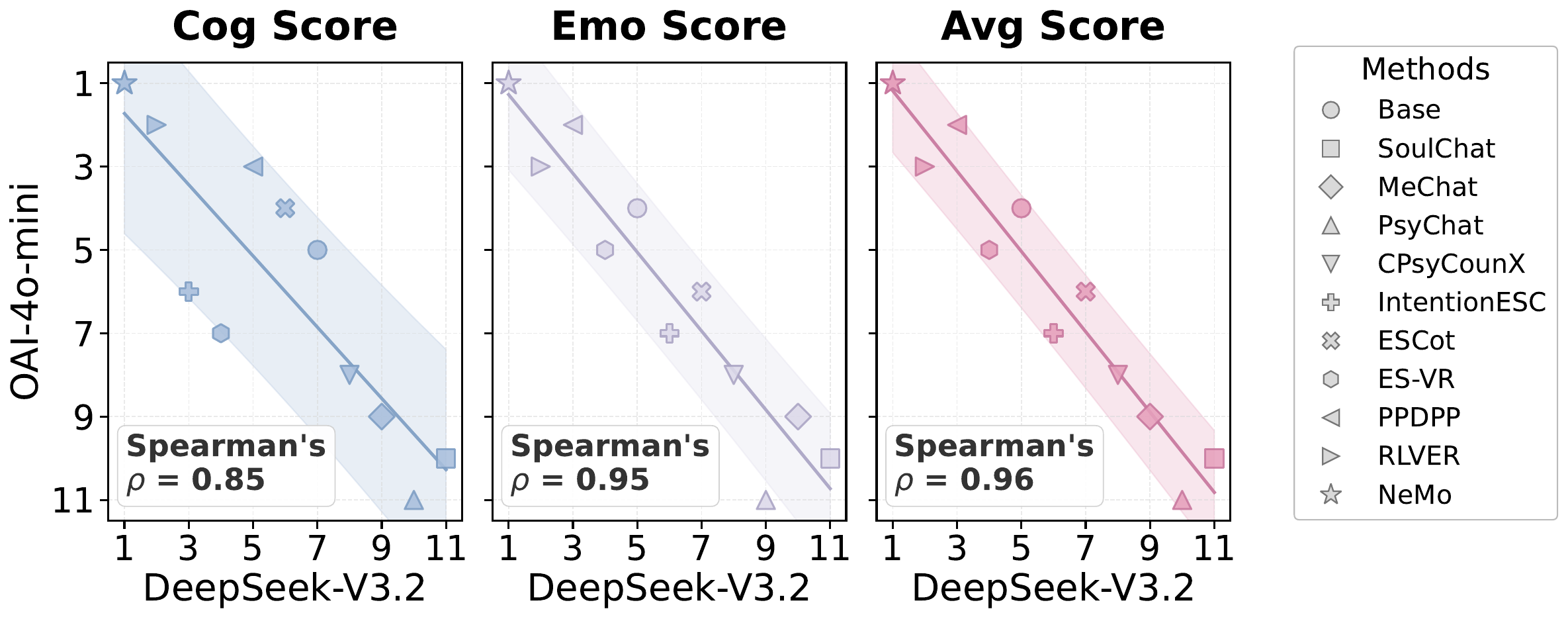}
        \caption{
        Cross-simulator rank consistency. We use \texttt{DeepSeek-v3.2} and \texttt{OAI-4o-mini} as simulator backbones and provide detailed results in Table~\ref{tab:simulator_comparison}.
        }
        \label{fig:rank_consistency}
    \end{minipage}
    \vspace{-0.5cm}
\end{figure*}

Since both training and evaluation rely on MindFlow, we verify its reliability in Figure~\ref{fig:rank_consistency} from two aspects. \textbf{(1) Human validation.} Annotators evaluate the reasonableness of key simulator outputs, including the preference vector $w_t$, update signal $\Delta_t$, user utterance $u_t$, and reward signal $r_t$. \textbf{(2) Cross-model consistency.} We replace the simulator's base model and find that the rankings of existing methods remain highly consistent, suggesting that MindFlow captures stable user-state transition patterns rather than model-specific preferences. More details are provided in Appendix~\ref{sec:human_validation_mindflow}.

\section{NeMo: Need-Aware Multi-Objective Policy Optimization}
\label{sec:optimization}

To improve policy learning under evolving user support preferences, we propose NeMo, a need-aware multi-objective reinforcement learning method for emotional support conversations. NeMo converts the multi-dimensional feedback from MindFlow into step-level training signals through two designs: multi-turn credit estimation for delayed user-state changes, and preference-guided reward aggregation for state-dependent objective priorities. Figure~\ref{fig:method_framework} provides an overview of our NeMo framework.

\subsection{Stepwise Reward Construction}
\label{sec:multi_turn_reward}

\textbf{Multi-Turn Reward Estimation.}
Although MindFlow provides user-side feedback at each turn, constructing reliable step-level rewards remains challenging. One-step differences offer cleaner attribution but may miss delayed user-state changes, whereas long-horizon returns capture delayed effects but can entangle the influence of later actions~\citep{ref:rl_introduction}. Therefore, we propose to estimate each response's local long-term effect through finite-horizon forward simulation. For a response $a_t$ at dialogue state $s_t$, we restore MindFlow to the latent mental state $M_t$, roll out future interactions to horizon $h$, and compute the multi-turn improvement reward:
\begin{equation}
\hat{r}_{t,h}
=
(\hat r_{t,h}^{emo},\hat r_{t,h}^{cog})
=
(r_{t+h}^{emo}-r_t^{emo},\; r_{t+h}^{cog}-r_t^{cog}).
\end{equation}
Here, $r_t$ denotes the simulator feedback, and $\hat{r}_{t,h}$ measures the user-state improvement associated with $a_t$ over horizon $h$. This design balances delayed-effect modeling with step-level credit assignment.

\textbf{Dynamic Reward Aggregation.}
During the conversation, the user's preference between emotional soothing and cognitive guidance evolves across turns. Thus, a single reward or static weighting may encourage generally helpful responses that mismatch the user's current support preference. To address this, we use the support preference vector $w_t=(w_t^{emo},w_t^{cog})$ derived by MindFlow to dynamically scalarize the multi-turn improvement reward $\hat{r}_{t,h}$:
\begin{equation}
\tilde r_{t}
=
\begin{cases}
w_t^{emo} \hat r_{t,h}^{emo}+w_t^{cog} \hat r_{t,h}^{cog},
& \text{if } \hat r_{t,h}^{emo} \hat r_{t,h}^{cog}\ge 0,\\
0.5\hat r_{t,h}^{emo}+0.5\hat r_{t,h}^{cog},
& \text{otherwise}.
\end{cases}
\label{eq:dynamic_reward}
\end{equation}
When the two reward dimensions improve or deteriorate together, the support preference vector emphasizes the objective that matches the user's current support preference and down-weights the other. When the two dimensions have opposite signs, we use uniform weighting instead, so that improvement in one dimension does not compensate for harm in the other. This reduces misleading credit assignment and encourages the policy to adapt to the user's current mental state.

\subsection{Need-Aware Policy Training}
\label{sec:grpo_learning}

After constructing the need-aware reward, we optimize the policy with GRPO~\citep{ref:grpo} at the step level. Given a training state $s_t$, the policy samples a group of candidate responses $\{a_t^i\}_{i=1}^{K}$. Each response is evaluated by MindFlow to obtain the scalar need-aware reward $\tilde r_t^i$, which is used for group-relative normalization. The policy is updated with the clipped objective:
\begin{equation}\small
\mathcal{L}(\theta)
=
-\mathbb{E}_{s_t,\{a_t^i\}}
\left[
\min
\left(
\rho_i A_i,\;
\mathrm{clip}(\rho_i,1-\epsilon,1+\epsilon)A_i
\right)
\right],
\rho_i=
\frac{\pi_{\theta}(a_t^i\mid s_t)}
{\pi_{\theta_{\mathrm{old}}}(a_t^i\mid s_t)},
A_i=
\frac{\tilde r_t^i-\mathrm{mean}(\{\tilde r_t^j\}_{j=1}^{K})}
{\mathrm{std}(\{\tilde r_t^j\}_{j=1}^{K})}.
\end{equation}
This objective favors responses that achieve higher need-aware rewards under the same dialogue state, encouraging the policy to generate responses aligned with the user's current support preference.

\section{Experiments}

\subsection{Experimental Setup}

\textbf{Datasets.}
We construct the main benchmark with MindFlow using static user personas from SAGE~\citep{ref:sage}. We assign each user an individualized arousal curve and remove cognition-style information from the original personas to avoid conflicts. MindFlow contains 500 training and 100 test profiles with matched train/test arousal-curve distributions. For generalization, we build MindFlow-Diverse with shifted arousal curves for Sim2Sim evaluation, and use PsyDial~\citep{ref:psydial} for Sim2Real evaluation. More details of the data are provided in Appendix~\ref{sec:mindflow_construction}.

\textbf{Metrics.} We evaluate both user-state improvement and response quality. For user-side metrics, MindFlow reports two state-improvement scores: Emo for emotional improvement and Cog for cognitive growth, and we report their average Avg as the overall user-state score. For response-level quality, we follow PsyDial~\citep{ref:psydial} and report nine supportive-skill scores on a 1--5 scale.

\textbf{Baselines.} 
We compare NeMo with four groups of baselines. 
(1) \textit{Vanilla Methods:} the base Qwen model~\citep{ref:qwen2.5} and four domain language models in ESC, including SoulChat~\citep{ref:soulchat}, MeChat~\citep{ref:smile}, PsyChat~\citep{ref:psychat}, and CPsyCounX~\citep{ref:cpsycoun}. 
(2) \textit{Supervised Methods:} IntentionESC~\citep{ref:intention_esc} and ESCot~\citep{ref:escot}, which augment dialogue data with chain-of-thought-style annotations to improve response quality and interpretability.
(3) \textit{Single-Reinforced Methods:} we include cognition-reinforced methods, such as ES-VR~\citep{ref:esvr}, which optimizes dialogue policy with user value feedback, and emotion-reinforced methods, such as PPDPP~\citep{ref:ppdpp} and RLVER~\citep{ref:rlver}, which use emotion-related signals as rewards.
(4) \textit{Multi-Reinforced Methods:} NeMo, which jointly optimizes emotional and cognitive objectives with need-aware stepwise reward.

\textbf{Implementation details.} 
We use \texttt{Qwen2.5-7B-Instruct} as the backbone policy and apply a lightweight SFT warm-up with LlamaFactory~\citep{ref:llamafactory} before GRPO. We follow RLVER~\citep{ref:rlver} for the inference prompt and cap dialogues at 10 turns. We roll out the base model on training profiles to obtain $\mathcal{D}_{\mathrm{roll}}=\{(s_t,a_t)\}$, use preference-matched response steps for SFT, and resample the unfiltered set to balance user support preferences for GRPO training. During GRPO, the policy samples candidate responses at each state $s_t$ and receives need-aware rewards from the simulator. We reserve 50 of the 500 training profiles for validation and use the final checkpoint for evaluation. GRPO is implemented with VeRL~\citep{ref:verl} using full-parameter fine-tuning, learning rate $1e$-$6$, batch size 32, group size 4, and 80 training steps. We tune the reward horizon $h\in\{1,2,3\}$ to balance immediate and long-term effects.

\subsection{Main Results}

\begin{table}[!t]
\centering
\caption{Main Performance on MindFlow. We highlight the \colorbox[HTML]{C4DBF1}{best}, the \colorbox[HTML]{DCEAF7}{second-best}, and the \colorbox[HTML]{F0F6FC}{third-best} results.}
\label{tab:main_results}
% Adjust row spacing
\renewcommand{\arraystretch}{1.1}
% Scale the table to fit the page width
\resizebox{0.95\linewidth}{!}{%
\begin{tabular}{l|ccc|ccccccccc}
\toprule
\multirow{2}{*}{\textbf{Model}} & \multicolumn{12}{c}{\textbf{MindFlow}} \\
\cmidrule(lr){2-4} \cmidrule(lr){5-13}
& \textbf{Cog} & \textbf{Emo} & \textbf{Avg} & \textbf{Act.} & \textbf{Emp.} & \textbf{NJud.} & \textbf{CRes.} & \textbf{Gui.} & \textbf{ESel.} & \textbf{ICla.} & \textbf{Ope.} & \textbf{Ove.} \\
\midrule
Base         & 56.69 & 61.12 & 58.91 & 2.09 & 1.92 & 4.12 & 1.39 & 1.34 & 1.54 & 1.41 & 1.13 & 2.28 \\
\midrule
SoulChat        & 44.06 & 14.86 & 29.46 & 1.62 & 1.66 & 2.54 & 1.26 & 1.15 & 1.40 & 1.16 & 1.09 & 1.76 \\
MeChat          & 46.95 & 23.41 & 35.18 & 2.06 & \cellcolor{thirdbest}2.02 & 3.28 & \cellcolor{best}1.48 & 1.40 & \cellcolor{best}1.77 & 1.44 & 1.34 & 2.12 \\
PsyChat         & 44.84 & 23.53 & 34.19 & 1.65 & 1.61 & 2.89 & 1.16 & 1.45 & 1.42 & 1.39 & \cellcolor{secondbest}1.45 & 1.67 \\
CPsyCounX       & 49.54 & 30.88 & 40.21 & 1.83 & 1.78 & 3.24 & \cellcolor{secondbest}1.47 & \cellcolor{thirdbest}1.47 & 1.64 & 1.44 & \cellcolor{thirdbest}1.37 & 1.97 \\
\midrule
IntentionESC    & \cellcolor{thirdbest}59.23 & 55.82 & 57.53 & 1.95 & 1.84 & 3.82 & 1.34 & 1.38 & 1.49 & 1.34 & 1.20 & 2.18 \\
ESCot           & 57.37 & 55.70 & 56.54 & 1.89 & 1.86 & 3.70 & 1.30 & 1.22 & 1.44 & 1.27 & 1.10 & 2.09 \\
\midrule
ES-VR           & 59.09 & 62.15 & 60.62 & 2.05 & 1.88 & 4.06 & 1.40 & 1.32 & 1.53 & 1.34 & 1.11 & 2.19 \\
\midrule
PPDPP           & 58.17 & \cellcolor{thirdbest}63.39 & \cellcolor{thirdbest}60.78 & \cellcolor{thirdbest}2.15 & 1.97 & \cellcolor{thirdbest}4.17 & \cellcolor{thirdbest}1.41 & 1.34 & 1.53 & \cellcolor{thirdbest}1.50 & 1.14 & \cellcolor{thirdbest}2.33 \\
RLVER           & \cellcolor{secondbest}59.81 & \cellcolor{secondbest}73.95 & \cellcolor{secondbest}66.88 & \cellcolor{secondbest}2.46 & \cellcolor{secondbest}2.16 & \cellcolor{secondbest}4.58 & \cellcolor{thirdbest}1.41 & \cellcolor{secondbest}1.50 & \cellcolor{secondbest}1.68 & \cellcolor{secondbest}1.53 & 1.24 & \cellcolor{secondbest}2.70 \\
\midrule
NeMo           & \cellcolor{best}69.31 & \cellcolor{best}88.86 & \cellcolor{best}79.09 & \cellcolor{best}2.75 & \cellcolor{best}2.20 & \cellcolor{best}4.70 & \cellcolor{thirdbest}1.41 & \cellcolor{best}1.88 & \cellcolor{thirdbest}1.65 & \cellcolor{best}1.87 & \cellcolor{best}1.46 & \cellcolor{best}2.74 \\
\bottomrule
\end{tabular}%
}
\vspace{-0.3cm}
\end{table}

\begin{figure*}[!t]
    \centering
    \setlength{\abovecaptionskip}{0.15cm}
    \begin{minipage}[b]{0.46\textwidth}
        \centering
        \includegraphics[width=\linewidth]{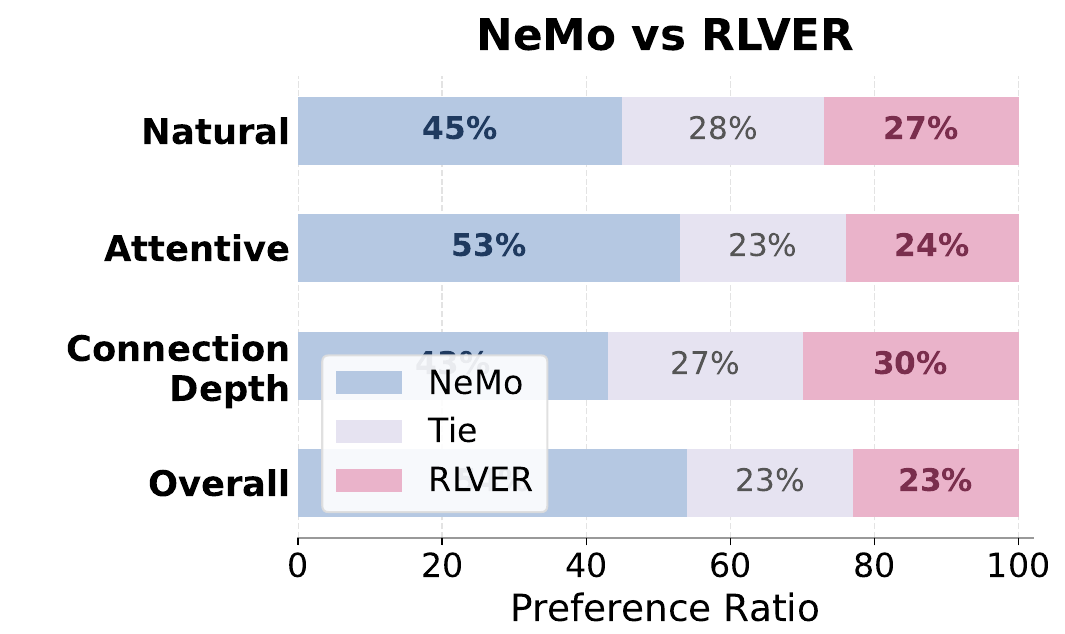}
        \caption{
        Human evaluation of response quality.
        }
        \label{fig:esc_human}
    \end{minipage}
    \hfill
    \begin{minipage}[b]{0.52\textwidth}
        \centering
        \includegraphics[width=\linewidth]{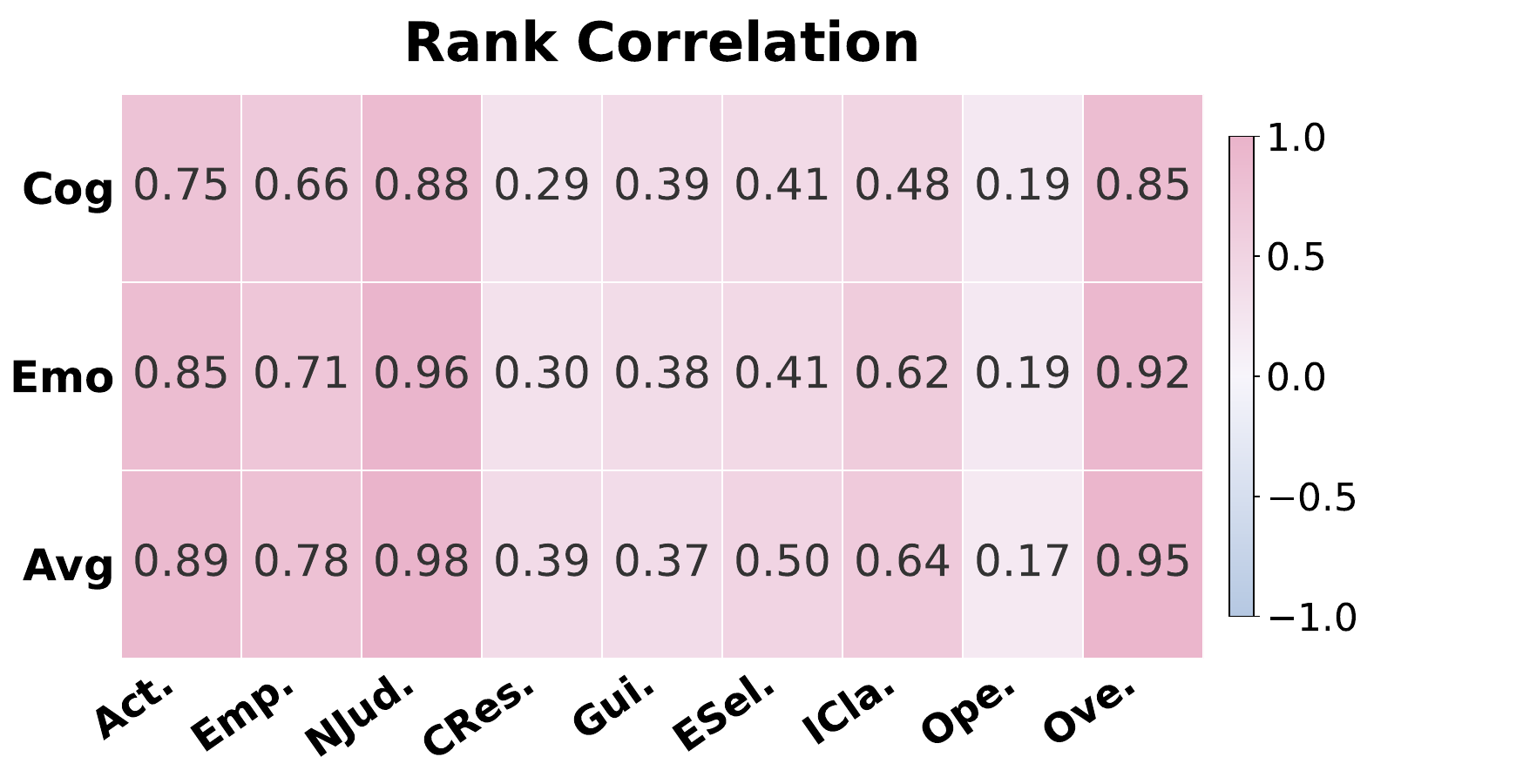}
        \caption{
        Correlation between user and textual evaluation.
        }
        \label{fig:esc_correlation}
    \end{minipage}
    \vspace{-0.4cm}
\end{figure*}

\textbf{How does NeMo compare with existing methods?}
We compare NeMo with existing emotional support dialogue methods on MindFlow and summarize three findings from Table~\ref{tab:main_results}.
(1) \textbf{User-centered reinforcement learning outperforms supervised methods.}
ES-VR, PPDPP, and RLVER achieve higher user-side Avg scores than IntentionESC and ESCot, indicating that optimizing user-state feedback is more effective for multi-turn state improvement than learning supportive response patterns from supervised data alone.
(2) \textbf{Need-aware multi-objective reinforcement learning further improves over single-objective reinforcement learning.}
Although RLVER is the strongest baseline with emotion-related rewards, NeMo improves Avg from 66.88 to 79.09, with absolute gains of 9.50 on Cog and 14.91 on Emo.
This suggests that emotional support requires jointly optimizing emotional relief and cognitive growth, rather than relying on a single user-state objective.
(3) \textbf{NeMo also improves response-level supportive quality.}
Besides user-side gains, NeMo achieves the best or near-best results on most response-level metrics, showing that need-aware optimization does not merely exploit simulator rewards but learns more effective supportive response behaviors.

\textbf{Do humans prefer trajectories generated by NeMo?}
We further conduct a trajectory-level human evaluation to examine whether NeMo's gains align with human judgments.
Specifically, we invite four annotators to compare complete dialogue trajectories from NeMo and the strongest baseline, RLVER, along four dimensions.
As shown in Figure~\ref{fig:esc_human}, NeMo is preferred on most dimensions, suggesting that need-aware optimization yields trajectories that are perceived as more attentive, more deeply connected, and more supportive overall. More details are provided in Appendix~\ref{sec:human_experiments}.

\textbf{Which supportive skills are associated with user-state improvement?}
We analyze correlations between user-side metrics and response-level supportive skills to examine how text-side behaviors relate to emotional relief and cognitive growth.
As shown in Figure~\ref{fig:esc_correlation}, Ove. shows the strongest correlation with Avg, indicating that overall supportive quality aligns closely with user-state improvement.
NJud., Act., and Emp. also correlate strongly with both Cog and Emo, suggesting that a non-judgmental, active, and empathetic context benefits both objectives.
In contrast, Ope. has relatively weak correlation with user-side metrics, while Gui. and ICla. are more related to cognitive support but remain timing-sensitive.
These results suggest that supportive skills are most effective when selected according to the user's current support preference, supporting our need-aware design.

\begin{table}[!t]
\centering
\caption{Ablation Study. We highlight the \colorbox[HTML]{F6DDE8}{most} and \colorbox[HTML]{FBEFF5}{second-most} effective modules.}
\label{tab:ablation}

\renewcommand{\arraystretch}{1.1}

\resizebox{0.97\linewidth}{!}{%
\begin{tabular}{l|ccc|ccccccccc}
\toprule
\multirow{2}{*}{\textbf{Model}} & \multicolumn{12}{c}{\textbf{User Simulator}} \\
\cmidrule(lr){2-4} \cmidrule(lr){5-13}
& \textbf{Cog} & \textbf{Emo} & \textbf{Avg} & \textbf{Act.} & \textbf{Emp.} & \textbf{NJud.} & \textbf{CRes.} & \textbf{Gui.} & \textbf{ESel.} & \textbf{ICla.} & \textbf{Ope.} & \textbf{Ove.} \\
\midrule
Our w/o $\mathcal{M}$ & \cellcolor{secondeffective}67.65 & \cellcolor{mosteffective}83.21 & \cellcolor{secondeffective}75.43 & \cellcolor{secondeffective}2.54 & \cellcolor{secondeffective}2.10 & \cellcolor{mosteffective}4.68 & \cellcolor{mosteffective}1.39 & \cellcolor{mosteffective}1.66 & \cellcolor{mosteffective}1.60 & \cellcolor{mosteffective}1.50 & \cellcolor{mosteffective}1.22 & \cellcolor{secondeffective}2.60 \\
Our w/o $\mathcal{D}$ & \cellcolor{mosteffective}65.12 & \cellcolor{secondeffective}83.89 & \cellcolor{mosteffective}74.51 & \cellcolor{mosteffective}2.50 & \cellcolor{mosteffective}2.04 & \cellcolor{mosteffective}4.68 & \cellcolor{secondeffective}1.41 & \cellcolor{secondeffective}1.89 & \cellcolor{secondeffective}1.62 & \cellcolor{secondeffective}1.65 & \cellcolor{secondeffective}1.48 & \cellcolor{mosteffective}2.59 \\
\midrule
NeMo & 69.31 & 88.86 & 79.09 & 2.75 & 2.20 & 4.70 & 1.41 & 1.88 & 1.65 & 1.87 & 1.46 & 2.74 \\
\bottomrule
\end{tabular}%
}
\end{table}

\subsection{Ablation Study}
\textbf{Are the key components of NeMo effective?}
We conduct ablation studies to examine two key designs in NeMo.
Specifically, \textit{NeMo (w/o $\mathcal{M}$)} replaces multi-turn rewards with adjacent-turn state differences, while \textit{NeMo (w/o $\mathcal{D}$)} replaces dynamic aggregation with fixed $0.5/0.5$ weighting over emotional and cognitive rewards.
As shown in Table~\ref{tab:ablation}, both variants degrade performance.
\textit{NeMo (w/o $\mathcal{M}$)} reduces Avg from 79.09 to 75.43 and Emo from 88.86 to 83.21, indicating that adjacent-turn differences miss delayed effects.
\textit{NeMo (w/o $\mathcal{D}$)} lowers Avg to 74.51, showing that fixed weights fail to capture changing support preferences.
Overall, NeMo achieves the best Cog, Emo, and Avg scores, demonstrating the complementarity of multi-turn reward estimation and dynamic reward aggregation.

\vspace{-0.2cm}
\subsection{Generalization Analysis}

\begin{table}[!t]
\centering
\caption{Generalization Performance. We highlight the \colorbox[HTML]{DDD3F2}{best}, the \colorbox[HTML]{EDE7F8}{second-best}, and the \colorbox[HTML]{F7F3FC}{third-best} results.}
\label{tab:generalization}

% Adjust row spacing
\renewcommand{\arraystretch}{1.1}

% Scale the table to fit the page width
\resizebox{0.95\linewidth}{!}{%
\begin{tabular}{l|ccc|ccccccccc}
\toprule
\multirow{2}{*}{\textbf{Model}} & \multicolumn{3}{c|}{\textbf{Sim2Sim}} & \multicolumn{9}{c}{\textbf{Sim2Real}} \\
\cmidrule(lr){2-4} \cmidrule(lr){5-13}
& \textbf{Cog} & \textbf{Emo} & \textbf{Avg} & \textbf{Act.} & \textbf{Emp.} & \textbf{NJud.} & \textbf{CRes.} & \textbf{Gui.} & \textbf{ESel.} & \textbf{ICla.} & \textbf{Ope.} & \textbf{Ove.} \\
\midrule
Base         & 55.27 & 47.96 & 51.62 & 1.39 & 1.81 & \cellcolor{gthirdbest}4.40 & 1.48 & 2.40 & 2.08 & 1.69 & 2.54 & 2.34 \\
\midrule
SoulChat        & 44.22 & 16.06 & 30.14 & 1.19 & 1.50 & 4.00 & 1.30 & 1.90 & 1.68 & 1.47 & 2.21 & 1.93 \\
MeChat          & 46.79 & 25.48 & 36.14 & 1.16 & 1.45 & 4.06 & 1.33 & 1.89 & 1.65 & 1.64 & 2.23 & 1.83 \\
PsyChat         & 46.11 & 29.10 & 37.61 & 1.31 & 1.58 & 4.21 & 1.22 & 1.53 & 1.40 & 1.44 & 1.75 & 1.68 \\
CPsyCounX       & 48.62 & 27.79 & 38.21 & 1.28 & 1.54 & 4.04 & 1.35 & 2.26 & 1.86 & \cellcolor{gthirdbest}1.78 & 2.47 & 1.93 \\
\midrule
IntentionESC    & 50.82 & 34.00 & 42.41 & \cellcolor{gthirdbest}1.62 & 1.90 & 4.31 & \cellcolor{gbest}1.75 & 2.35 & \cellcolor{gthirdbest}2.14 & 1.47 & 1.95 & \cellcolor{gthirdbest}2.74 \\
ESCot           & 50.81 & 31.99 & 41.40 & 1.56 & \cellcolor{gthirdbest}1.96 & 4.22 & \cellcolor{gsecondbest}1.62 & 1.81 & 1.85 & 1.28 & 1.56 & 2.64 \\
\midrule
ES-VR           & 55.04 & 50.52 & 52.78 & 1.40 & 1.73 & 4.28 & 1.47 & 2.38 & 2.03 & 1.72 & 2.57 & 2.27 \\
\midrule
PPDPP           & \cellcolor{gthirdbest}55.74 & \cellcolor{gthirdbest}51.86 & \cellcolor{gthirdbest}53.80 & 1.37 & 1.75 & 4.30 & 1.44 & 2.36 & 2.01 & 1.73 & \cellcolor{gthirdbest}2.60 & 2.26 \\
RLVER           & \cellcolor{gsecondbest}64.93 & \cellcolor{gsecondbest}74.76 & \cellcolor{gsecondbest}69.85 & \cellcolor{gsecondbest}1.92 & \cellcolor{gsecondbest}2.26 & \cellcolor{gsecondbest}4.65 & 1.46 & \cellcolor{gsecondbest}2.65 & \cellcolor{gbest}2.49 & \cellcolor{gbest}1.88 & \cellcolor{gbest}2.79 & \cellcolor{gsecondbest}2.96 \\
\midrule
NeMo            & \cellcolor{gbest}67.55 & \cellcolor{gbest}79.14 & \cellcolor{gbest}73.35 & \cellcolor{gbest}2.29 & \cellcolor{gbest}2.35 & \cellcolor{gbest}4.73 & \cellcolor{gthirdbest}1.52 & \cellcolor{gbest}3.00 & \cellcolor{gsecondbest}2.45 & \cellcolor{gsecondbest}1.83 & \cellcolor{gsecondbest}2.72 & \cellcolor{gbest}3.11 \\
\bottomrule
\end{tabular}%
}
\end{table}

\begin{figure*}[!t]
    \centering
    \setlength{\abovecaptionskip}{0.15cm}
    \begin{minipage}[b]{0.275\textwidth}
        \centering
        \includegraphics[width=\linewidth]{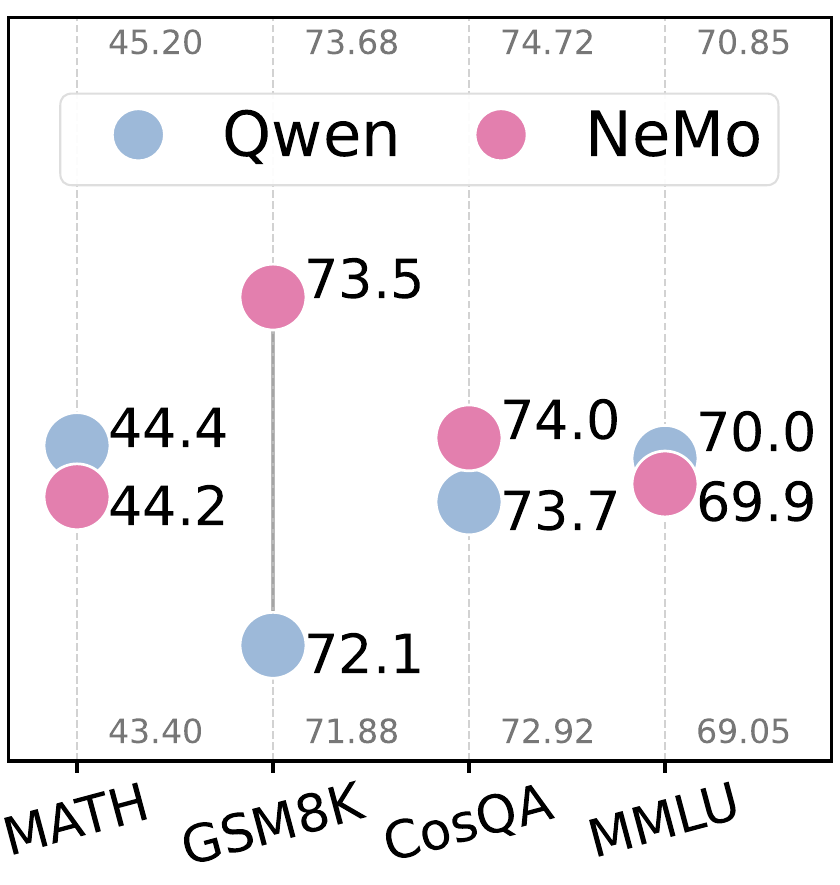}
        \caption{
        Preservation of general reasoning capability.
        }
        \label{fig:general_reasoning}
    \end{minipage}
    \hfill
    \begin{minipage}[b]{0.675\textwidth}
        \centering
        \includegraphics[width=\linewidth]{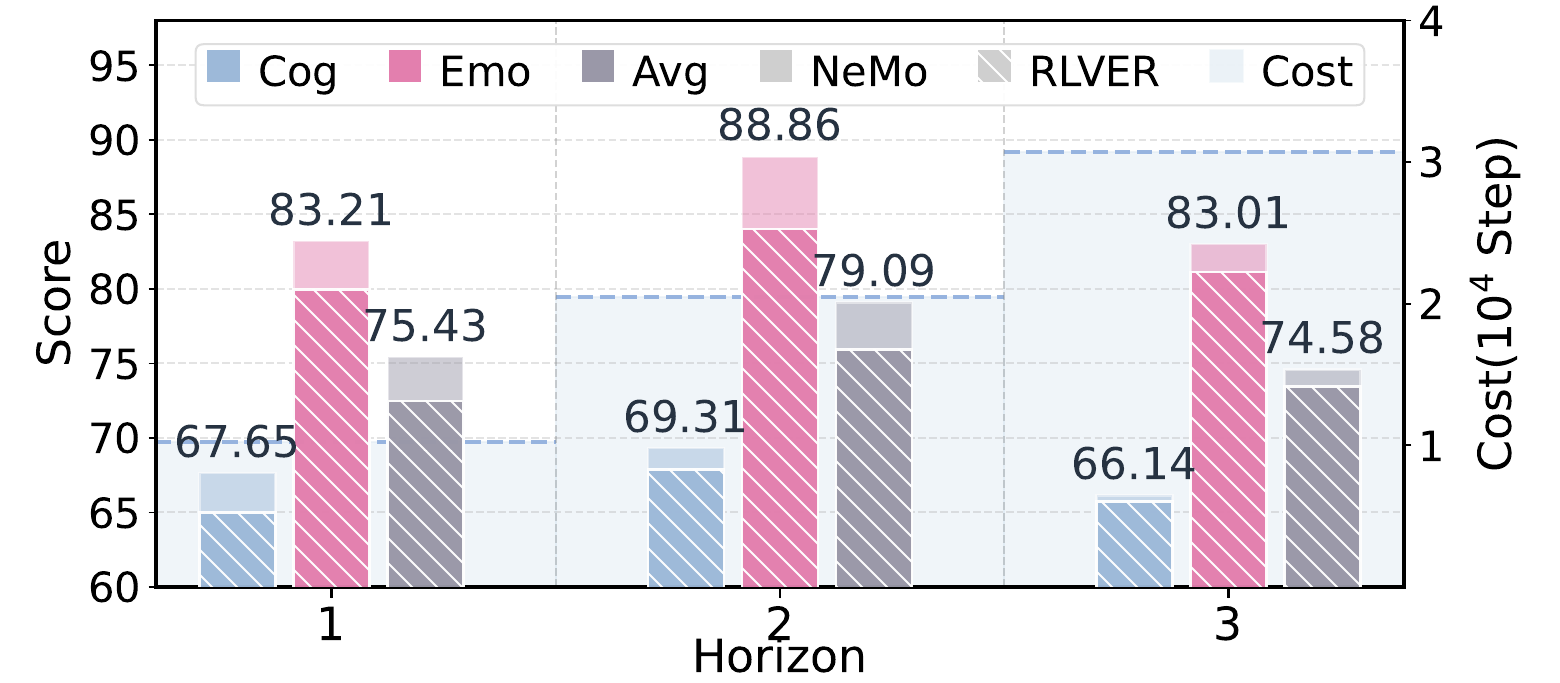}
        \caption{
        Effect of multi-turn reward estimation. We evaluate the design from both effectiveness and efficiency perspectives.
        }
        \label{fig:multi_turn_reward}
    \end{minipage}

    \vspace{-0.4cm} 
\end{figure*}

\textbf{Can NeMo generalize to shifted simulated users and real-world dialogues?}
We further evaluate the generalization ability of NeMo under two settings: Sim2Sim, where MindFlow-Diverse uses shifted arousal-curve configurations, and Sim2Real, where PsyDial~\citep{ref:psydial} is used as a real-world dialogue benchmark.
As shown in Table~\ref{tab:generalization}, NeMo maintains the best overall performance in both settings.
On Sim2Sim, NeMo achieves 73.35 Avg, outperforming the strongest baseline RLVER with 69.85 Avg.
This indicates that NeMo remains effective even when the user need-shift patterns differ from the training distribution.
On Sim2Real, NeMo also achieves the best or near-best results on most response-level skills, suggesting that the need-aware policy learned from simulation can transfer to real dialogue data rather than overfitting to a specific simulator.
Overall, these results show that NeMo learns robust and generalizable support strategies after training.

\textbf{Does NeMo preserve general reasoning ability?}
Since NeMo is trained in full-parameter settings on emotional support dialogues, we further examine whether this process harms the policy's general abilities.
Specifically, we compare the Base model and our trained model on general benchmarks, including MATH~\citep{ref:math}, GSM8K~\citep{ref:gsm8k}, CosmosQA~\citep{ref:cosmosqa}, and MMLU~\citep{ref:mmlu}.
As shown in Figure~\ref{fig:general_reasoning}, the trained model maintains performance close to the Base model on most tasks, without clear degradation.
This suggests that NeMo's emotional support optimization does not substantially damage the base model's original mathematical reasoning or commonsense question-answering abilities.
NeMo mainly learns better strategies for emotional support scenarios, rather than overfitting at the cost of general capabilities.

\subsection{Effectiveness of Multi-Turn Reward Estimation}
\textbf{How does multi-turn reward estimation affect training?}
Here, we first study the effect of the forward simulation horizon $h$.
From Figure~\ref{fig:multi_turn_reward}, NeMo performs best when $h=2$.
A shorter horizon mainly captures adjacent-turn changes and may miss delayed user-state effects, while a longer horizon introduces noise from later responses and state transitions, making credit assignment less stable.
Thus, $h=2$ provides a better trade-off between delayed-effect modeling and reward noise.

We further compare NeMo's step-level rollout with trajectory-level rollout under the same dialogue-turn budget.
Trajectory-level methods such as RLVER~\citep{ref:rlver} sample complete dialogue trajectories, compute group-relative advantages from final returns, and assign the same advantage to all steps.
This captures long-horizon outcomes but obscures each response's contribution.
In contrast, NeMo restores MindFlow at each state and rolls out from the current response to estimate its local long-term effect on future user states.
The stronger performance of step-level rollout shows that multi-turn reward improves training through more accurate step-level credit assignment.

\subsection{Effectiveness of Dynamic Reward Aggregation}

\begin{figure*}[!t]
    \centering
    \setlength{\abovecaptionskip}{0.15cm}
    \begin{subfigure}[b]{0.50\textwidth}
        \centering
        \includegraphics[width=\linewidth]{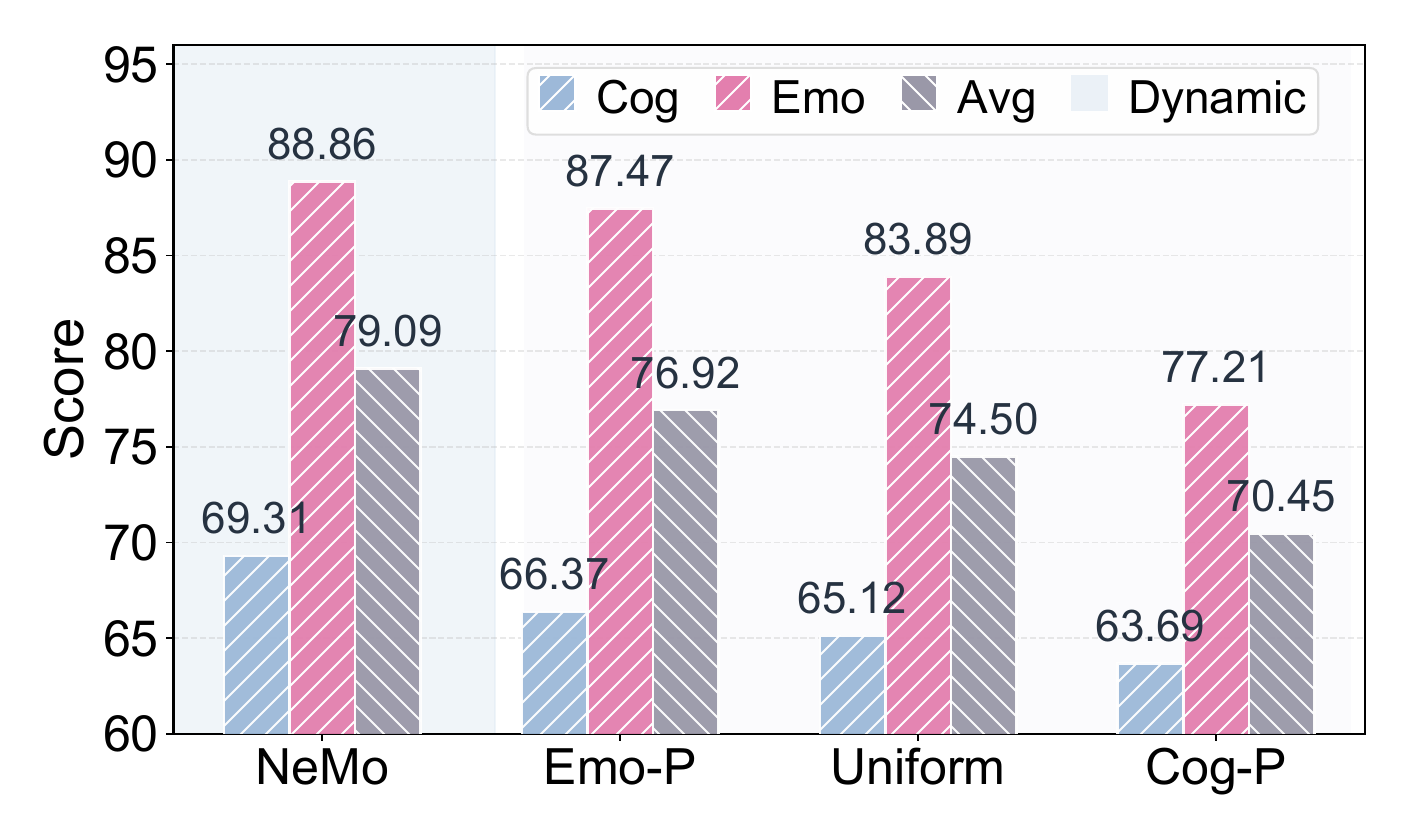}
        \label{fig:dynamic_user}
    \end{subfigure}
    \hfill
    \begin{subfigure}[b]{0.49\textwidth}
        \centering
        \includegraphics[width=\linewidth]{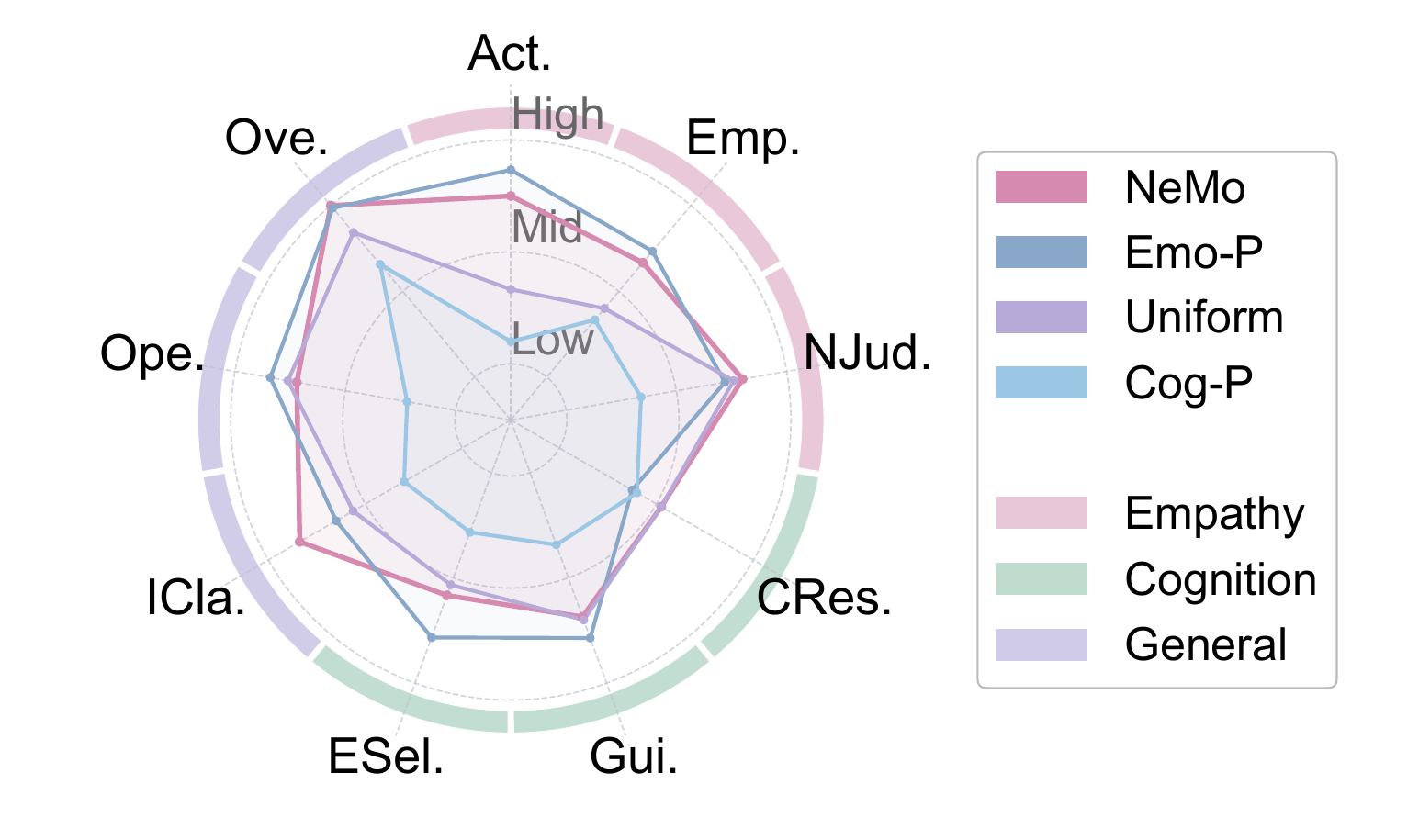}
        \label{fig:dynamic_text}
    \end{subfigure}
    \vspace{-0.8cm}
    \caption{Effectiveness of dynamic reward aggregation.}
    \label{fig:dynamic_reward}
\vspace{-0.4cm}
\end{figure*}

\textbf{How does dynamic reward aggregation affect optimization?}
We compare NeMo with three fixed-weight variants: \textit{Emo-P} sets $w_t^{cog}=0$ and optimizes only emotional rewards, \textit{Cog-P} sets $w_t^{cog}=1$ and optimizes only cognitive rewards, and \textit{Uniform} sets $w_t^{cog}=0.5$ to equally weight both objectives.
In contrast, NeMo uses the dynamic aggregation in Eq.~\ref{eq:dynamic_reward}, where $w_t=(w_t^{emo}, w_t^{cog})$ is derived from the user's current support preference.
As shown in the left panel of Figure~\ref{fig:dynamic_reward}, NeMo achieves the best Cog, Emo, and Avg scores, showing the advantage of dynamic aggregation over fixed weighting.
Performance also improves from \textit{Cog-P} to \textit{Uniform} and then to \textit{Emo-P} as $w_t^{emo}$ increases, suggesting that early-stage users often benefit more from empathy and emotional acceptance. By dynamically balancing emotional relief and cognitive growth, NeMo better matches evolving user preferences.

We further evaluate response-level supportive skills to examine how different reward designs shape the model's support behaviors.
In right panel of Figure~\ref{fig:dynamic_reward}, fixed rewards induce different skill biases: \textit{Emo-P} performs better on affective and general skills such as Emp., NJud., and Ove., but is weaker on cognition-related skills.
In contrast, NeMo yields a more balanced skill profile, preserving empathy and non-judgment while maintaining clarification and guidance.
This suggests that the dynamic aggregation design can help agents adapt support behaviors to the user's current mental state.

\vspace{-0.2cm}
\section{Related Work}
\vspace{-0.2cm}
\textbf{User-Centered Emotional Support Conversations.}
Emotional support conversations aim to improve users' subjective experience and psychological states, rather than merely response fluency or surface-level support. Early studies mainly learn supportive behaviors from real-world~\citep{ref:esconv,ref:xinling,ref:psyqa} or LLM-synthesized dialogues~\citep{ref:aug_esc,ref:extes,ref:chatcounselor}, often enriched with longer contexts~\citep{ref:smile}, user profiles~\citep{ref:socialsim,ref:aug_esc,ref:extes}, or intermediate intentions and reasoning~\citep{ref:escot,ref:intention_esc,ref:compeer}. Although these data-centric methods provide useful behavioral priors, their supervision remains largely text-side. Recent simulator-based methods incorporate user-state changes into training and evaluation through emotional~\citep{ref:rlver,ref:ppdpp} or cognitive~\citep{ref:esvr} rewards. However, these methods typically optimize a single user-state objective. In contrast, we formulate emotional support as a need-aware multi-objective problem that dynamically balances emotional and cognitive objectives according to users' evolving needs.

\textbf{User Simulation for Emotional Support.}
User simulators provide a scalable way to approximate user-side reactions in ESC.
Early simulators were trained on real or LLM-augmented human dialogues~\citep{ref:esc_eval}, but the limited scale and scenario coverage of such data make the resulting simulators mainly suitable for simple dialogue patterns, while the lack of explicit user-state signals forces their training to still rely largely on text-side supervision.
Recent work has turned to LLM-based simulators, either by designing diverse personas~\citep{ref:socialsim,ref:esvr,ref:personas} or by introducing psychological theories to model user-state transitions~\citep{ref:patient_psi,ref:sage}.
However, existing simulators still lack explicit modeling of how multiple support needs coexist and evolve across turns.
We address this by modeling user-need transitions with dual-system theory and using personalized arousal curves to maintain consistent cognitive characteristics, providing multi-dimensional and psychologically grounded feedback signals.

\vspace{-0.2cm}
\section{Conclusion}
\vspace{-0.2cm}
In this work, we formulate emotional support dialogue as a need-aware multi-objective decision-making problem, where agents should dynamically balance emotional relief and cognitive growth according to users' evolving mental needs. We introduced MindFlow, a psychologically grounded user simulator that explicitly models emotional and cognitive states and derives state-dependent support preferences, and proposed NeMo, a reinforcement learning framework that combines forward simulation with dynamic reward aggregation for need-aware policy optimization. Experiments show that NeMo improves both user-side state metrics and response-level supportive quality, and further demonstrates robust generalization under both Sim2Sim and Sim2Real evaluations.

\section*{Acknowledgements}
This work is supported in part by Huawei Innovation Research Programs. We gratefully acknowledge the support from MindSpore, CANN, and Ascend AI Processor used for this research.

\bibliographystyle{plainnat}
\bibliography{reference}

\newpage
\appendix
\etocdepthtag.toc{appendix}
\begingroup
\etocsettagdepth{main}{none}
\etocsettagdepth{appendix}{subsection}
\etocsettocstyle{\section*{Appendix Contents}}{}
\tableofcontents
\endgroup
\clearpage

\section{Limitations}
While MindFlow and NeMo present a promising framework for emotionally intelligent LLM agent construction, there are still several limitations for future work. First, both training and evaluation still rely substantially on the MindFlow simulator. Although we validate the simulator's reliability through human annotations and cross-model consistency experiments, simulator-based feedback remains an approximation of real user reactions, which may not fully cover the more complex, nuanced, and highly individualized psychological changes in real emotional support scenarios. Second, this work mainly formulates emotional support as a bi-objective problem involving emotional relief and cognitive growth. Real support needs may involve richer user-side dimensions and may also benefit from additional text-side indicators. Incorporating these broader signals could provide more comprehensive supervision and further improve need-aware policy optimization.

\section{Ethical Considerations}
This work focuses on emotional support dialogue and aims to improve LLM agents’ ability to respond to users’ evolving emotional and cognitive needs. Since this setting involves users’ mental states and potentially vulnerable situations, our method should not be regarded as a substitute for professional counseling, clinical diagnosis, or therapeutic intervention. In deployment, systems should clearly communicate their non-clinical nature and redirect users to professional or emergency resources when risks such as self-harm, harm to others, or crisis situations are detected. Our training and evaluation are mainly based on user simulation and existing datasets, rather than direct intervention with real users, which reduces potential psychological risks during experimentation. However, simulators and datasets may still contain biases and may not fully reflect diverse user needs across cultures, ages, languages, and mental health conditions. Therefore, real-world use should involve stricter safety evaluation, fairness analysis, privacy protection, and human oversight.

\section{Notation Summary}

Table~\ref{tab:notations} summarizes the key notations used in MindFlow simulation and NeMo policy training.

\begin{table}[!ht]
\vspace{-6pt}
\centering
\caption{Summary of key notations.}
\label{tab:notations}
\small
\renewcommand{\arraystretch}{1.2}
\begin{tabular}{l p{0.78\linewidth}}
\toprule
\textbf{Symbol} & \textbf{Description} \\
\midrule
$M_t$ & User latent state at turn $t$, where $M_t = (e_t, c_t)$ \\
$e_t$ & Emotional state represented by valence and arousal \\
$c_t$ & Cognitive state in natural language, including appraisal and planning \\
$w_t$ & Support preference vector, where $w_t = (w_t^{\mathrm{emo}}, w_t^{\mathrm{cog}})$ \\
$\tilde{\Delta}_t$ & Coarse state-update signal from fast thinking, where $\tilde{\Delta}_t = (\tilde{\Delta}_t^{\mathrm{emo}}, \tilde{\Delta}_t^{\mathrm{cog}})$ \\
$\Delta_t$ & Final state-update signal from slow thinking, where $\Delta_t = (\Delta_t^{\mathrm{emo}}, \Delta_t^{\mathrm{cog}})$ \\
$r_t$ & Multi-objective feedback, where $r_t = (r_t^{\mathrm{emo}}, r_t^{\mathrm{cog}})$ \\
$\hat{r}_{t,h}$ & Multi-objective improvement over horizon $h$, where $\hat{r}_{t,h} = (\hat{r}_{t,h}^{\mathrm{emo}}, \hat{r}_{t,h}^{\mathrm{cog}})$ \\
$\tilde{r}_t$ & Aggregated scalar reward for policy optimization \\
\bottomrule
\end{tabular}
\vspace{-6pt}
\end{table}

\section{MindFlow Construction Details}
\label{sec:mindflow_construction_details}

\subsection{Definitions of Emotional and Cognitive States}
\label{sec:state_definitions}

MindFlow decomposes the user's latent mental state into emotional and cognitive states. Following Russell's circumplex model, the emotional state is a two-dimensional continuous vector of valence and arousal. Valence measures emotional polarity in \((-1,1)\), where larger values indicate more positive affect and smaller values indicate more negative affect. Arousal measures activation, tension, or engagement in \((0,1)\), where larger values indicate a more activated or tense state. The cognitive state is represented in natural language to keep the user's internal reasoning interpretable. It contains two fields: appraisal and solution. Appraisal describes how the user understands the current difficulty, attributes causes, and interprets their own and others' behaviors. Solution describes the user's current coping plan, including action direction, executable steps, and perceived control. Together, these states allow MindFlow to track both affective changes and cognitive progress across turns.

\subsection{Construction of MindFlow and MindFlow-Diverse}
\label{sec:mindflow_construction}

\begin{figure}[!t]
    \centering
    \setlength{\abovecaptionskip}{0.15cm}
    \includegraphics[width=0.9\linewidth]{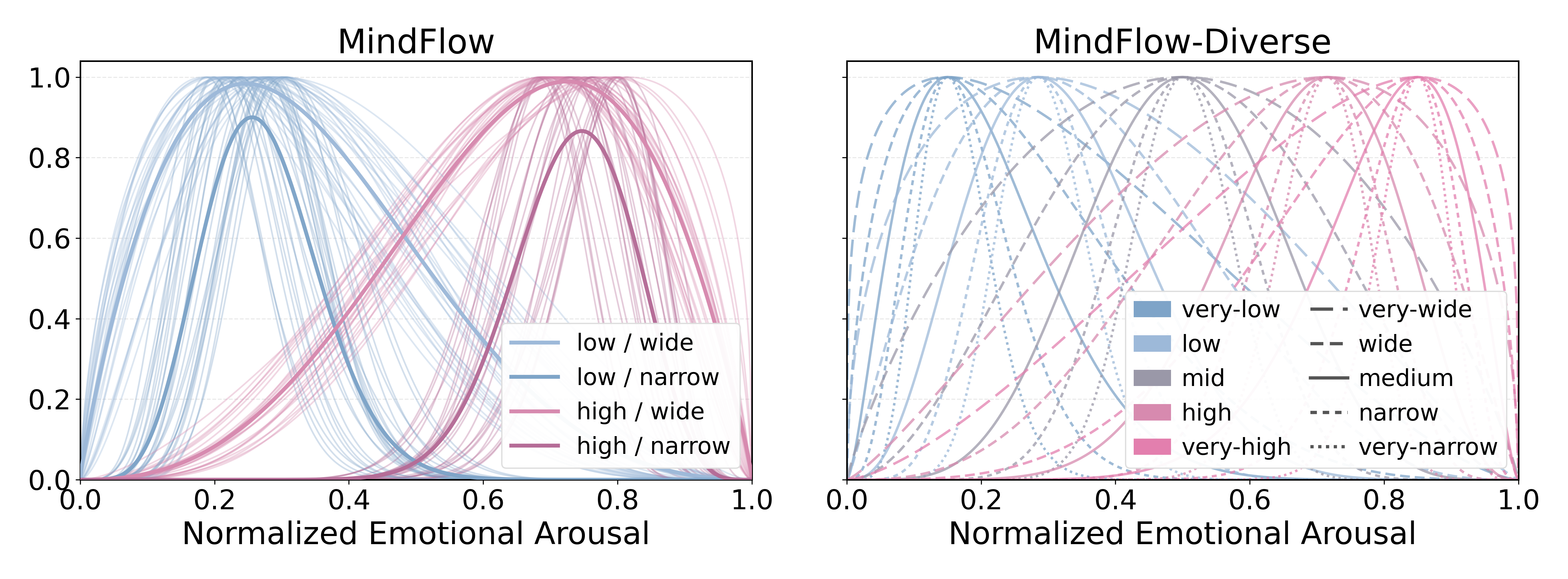}
    \caption{
    User arousal-curve distributions in the MindFlow and MindFlow-Diverse test sets.
    }
    \label{fig:arousal_curve_distribution}
\end{figure}

For data construction, we use the static user profiles and scenarios from SAGE~\citep{ref:sage}, and introduce \texttt{DeepSeek-v3.2} to remove descriptions of users' cognitive characteristics from the profiles, avoiding conflicts between static personas and the arousal curves that control later dialogue dynamics. To construct personalized arousal curves, we first sample two curve parameters, \(a_u\) and \(d_u\), where \(a_u\) controls the user's optimal arousal location for cognitive support and \(d_u\) controls the curve width. Intuitively, \(a_u\) determines when a user becomes most receptive to cognitive guidance, while \(d_u\) reflects how sensitive the user is to deviations from this optimal state. We then convert them into the Beta-distribution parameters \(\alpha_u = 1 + a_u\cdot d_u\) and \(\beta_u = 1 + (1-a_u)\cdot d_u\). In implementation, we constrain the sampling ranges of \(a_u\) and \(d_u\) to ensure that the resulting Beta distribution forms a unimodal inverted-U curve. At each dialogue turn, MindFlow inputs the user's normalized emotional arousal into this curve and uses the normalized density as the cognitive-support weight, yielding state-dependent preferences between emotional and cognitive support. We use different sampling ranges of \(a_u\) and \(d_u\) to construct the arousal curves of MindFlow and MindFlow-Diverse, and Figure~\ref{fig:arousal_curve_distribution} visualizes the sampled curves, showing the distributional difference between the two settings.

\subsection{Efficiency Analysis of MindFlow.}

We compare the inference cost of MindFlow with the SAGE simulator under the same evaluation setting. Specifically, we use both simulators to evaluate \texttt{DeepSeek-v3.2} on the same 100 MindFlow test profiles, and measure the token cost required by the user simulator to generate one dialogue turn. The SAGE~\citep{ref:sage} simulator requires 1551.5 tokens per turn on average, whereas MindFlow requires 713.7 tokens per turn. This corresponds to a 54.0\% reduction in per-turn inference tokens. This gain mainly comes from MindFlow's structured state representation and compact transition process: instead of repeatedly prompting the simulator to infer all latent factors from the full dialogue context, MindFlow explicitly maintains emotional and cognitive states and updates them through the fast--slow transition mechanism. As a result, MindFlow provides more efficient user-state simulation while preserving the key information needed for need-aware feedback.

\subsection{Details of Cross-Model Consistency}

We examine the robustness of MindFlow to the choice of backbone model by evaluating the same set of methods with \texttt{DeepSeek-v3.2} and \texttt{OAI-4o-mini} as simulator backbones. As shown in Table~\ref{tab:simulator_comparison} and Figure~\ref{fig:rank_consistency}, the rankings of existing methods remain highly consistent across the two backbones, despite differences in absolute scores. This consistency suggests that MindFlow provides stable comparative evaluations across different backbone models.

\begin{table}[t]
\centering
\caption{Detailed results of cross-simulator consistency in Figure~\ref{fig:rank_consistency}.}
\label{tab:simulator_comparison}
\renewcommand{\arraystretch}{1.1}
\resizebox{0.59\linewidth}{!}{%
\begin{tabular}{l|ccc|ccc}
\toprule
\multirow{2}{*}{\textbf{Model}} 
& \multicolumn{3}{c|}{\textbf{DeepSeek-v3.2}} 
& \multicolumn{3}{c}{\textbf{OAI-4o-mini}} \\
\cmidrule(lr){2-4} \cmidrule(lr){5-7}
& \textbf{Cog} & \textbf{Emo} & \textbf{Avg} 
& \textbf{Cog} & \textbf{Emo} & \textbf{Avg} \\
\midrule
Base      & 56.69 & 61.12 & 58.91 & 71.10 & 62.70 & 66.90 \\
\midrule
SoulChat     & 44.06 & 14.86 & 29.46 & 51.25 & 24.47 & 37.86 \\
MeChat       & 46.95 & 23.41 & 35.18 & 56.55 & 38.85 & 47.70 \\
PsyChat      & 44.84 & 23.53 & 34.19 & 47.50 & 23.50 & 35.50 \\
CPsyCounX    & 49.54 & 30.88 & 40.21 & 59.15 & 39.98 & 49.57 \\
\midrule
IntentionESC & 59.23 & 55.82 & 57.53 & 70.90 & 56.65 & 63.78 \\
ESCot        & 57.37 & 55.70 & 56.54 & 71.73 & 58.30 & 65.02 \\
\midrule
ES-VR        & 59.09 & 62.15 & 60.62 & 70.40 & 61.40 & 65.90 \\
\midrule
PPDPP        & 58.17 & 63.39 & 60.78 
             & 71.80 & 66.40 & 69.10 \\
RLVER        & 59.81 & 73.95 & 66.88 
             & 72.25 & 65.70 & 68.98 \\
\midrule
NeMo        & 69.31 & 88.86 & 79.09 
             & 72.86 & 88.28 & 80.57 \\
\bottomrule
\end{tabular}%
}
\end{table}

\subsection{Details of Human Validation}
\label{sec:human_validation_mindflow}
We construct a stratified annotation set of 100 single-step simulator transitions, balancing dialogue source, support-priority type, and turn position. Given the user profile, event background, dialogue history, previous mental state $M_t$, and system response $a_t$, annotators rate the reasonableness of the simulator-produced support priority $w_t$, state update $\Delta_t$, and next user utterance $u_{t+1}$, and the health reward $r_t$ on a 1--5 reasonableness scale. To avoid exposing methodological assumptions, reward values and arousal-curve parameters are hidden from annotators. Four college-student annotators label all transitions, and we treat scores no smaller than 3 as reasonable to compute the module-level validation rates reported in Figure~\ref{fig:eval_results}. The results show that MindFlow produces generally reasonable support priorities, state transitions, and user responses.

\section{Leaderboard Results}
\label{sec:leaderboard_results}

\subsection{Standard Evaluation}

\begin{table}[p]
\centering
\caption{Leaderboard on MindFlow (10 turns).}
\label{tab:mindflow_leaderboard}

\renewcommand{\arraystretch}{1.1}
\resizebox{0.6\linewidth}{!}{%
\begin{tabular}{c|l|ccc}
\toprule
\textbf{Rank} & \textbf{Model} & \textbf{Cog} & \textbf{Emo} & \textbf{Avg} \\
\midrule
1  & \texttt{oai-5-chat-latest}      & 73.86 & 88.36 & 81.11 \\
2  & \texttt{qwen3.5-plus}           & 75.36 & 86.10 & 80.73 \\
3  & \texttt{oai-4o}                 & 72.86 & 88.28 & 80.57 \\
4  & \texttt{qwen3-235b-a22b}        & 74.97 & 85.40 & 80.19 \\
5  & \texttt{gdm-3-flash-preview} & 73.61 & 86.25 & 79.93 \\
\rowcolor[HTML]{DCEAF7}
6  & NeMo                            & 69.31 & 88.86 & 79.09 \\
7  & \texttt{deepseek-v3}            & 73.17 & 84.23 & 78.70 \\
8  & NeMo-SFT                        & 64.42 & 82.55 & 73.48 \\
9  & \texttt{gdm-2.5-flash}       & 67.18 & 69.89 & 68.54 \\
10 & RLVER                           & 59.81 & 73.95 & 66.88 \\
11 & PPDPP                           & 58.17 & 63.39 & 60.78 \\
12 & ES-VR                           & 59.09 & 62.15 & 60.62 \\
13 & \texttt{deepseek-v3.2}          & 58.28 & 62.84 & 60.56 \\
14 & \texttt{qwen2.5-7b-instruct}    & 56.69 & 61.12 & 58.91 \\
15 & IntentionESC                    & 59.23 & 55.82 & 57.53 \\
16 & ESCot                           & 57.37 & 55.70 & 56.54 \\
17 & CPsyCounX                       & 49.54 & 30.88 & 40.21 \\
18 & MeChat                          & 46.95 & 23.41 & 35.18 \\
19 & PsyChat                         & 44.84 & 23.53 & 34.19 \\
20 & SoulChat                        & 44.06 & 14.86 & 29.46 \\
\bottomrule
\end{tabular}%
}

\vspace{12pt}

\centering
\caption{Leaderboard on MindFlow-Diverse (10 turns).}
\label{tab:mindflow_diverse_leaderboard}
\renewcommand{\arraystretch}{1.1}
\resizebox{0.6\linewidth}{!}{%
\begin{tabular}{c|l|ccc}
\toprule
\textbf{Rank} & \textbf{Model} & \textbf{Cog} & \textbf{Emo} & \textbf{Avg} \\
\midrule
1  & \texttt{oai-5-chat-latest}      & 74.19 & 87.46 & 80.82 \\
2  & \texttt{qwen3.5-plus}           & 75.80 & 85.37 & 80.58 \\
3  & \texttt{gdm-3-flash-preview} & 75.03 & 85.76 & 80.40 \\
4  & \texttt{oai-4o}                 & 73.24 & 86.45 & 79.84 \\
5  & \texttt{qwen3-235b-a22b}        & 75.32 & 83.79 & 79.56 \\
6  & \texttt{deepseek-v3}            & 72.75 & 80.51 & 76.63 \\
\rowcolor[HTML]{DCEAF7}
7  & NeMo                            & 67.55 & 79.14 & 73.35 \\
8  & NeMo-SFT                        & 65.29 & 77.97 & 71.63 \\
9  & RLVER                           & 64.93 & 74.76 & 69.85 \\
10 & \texttt{gdm-2.5-flash}       & 66.40 & 65.73 & 66.06 \\
11 & \texttt{deepseek-v3.2}          & 62.66 & 63.96 & 63.31 \\
12 & PPDPP                           & 55.74 & 51.86 & 53.80 \\
13 & ES-VR                           & 55.04 & 50.52 & 52.78 \\
14 & \texttt{qwen2.5-7b-instruct}    & 55.27 & 47.96 & 51.62 \\
15 & IntentionESC                    & 50.82 & 34.00 & 42.41 \\
16 & ESCot                           & 50.81 & 31.99 & 41.40 \\
17 & CPsyCounX                       & 48.62 & 27.79 & 38.21 \\
18 & PsyChat                         & 46.11 & 29.10 & 37.61 \\
19 & MeChat                          & 46.79 & 25.48 & 36.14 \\
20 & SoulChat                        & 44.22 & 16.06 & 30.14 \\
\bottomrule
\end{tabular}%
}
\end{table}

Tables~\ref{tab:mindflow_leaderboard} and~\ref{tab:mindflow_diverse_leaderboard} provide a broader comparison between NeMo, frontier models, and specialized emotional support systems. All proprietary models are evaluated through their official APIs using the versions listed in the tables. On MindFlow, NeMo ranks sixth overall with an Avg score of 79.09, improving over NeMo-SFT and RLVER by 5.61 and 12.21 points, respectively. These gains highlight the benefits of need-aware reinforcement learning beyond supervised warm-up and emotion-oriented optimization. Although several frontier models achieve higher Avg scores, NeMo remains competitive despite its much smaller backbone and obtains the highest Emo score, exceeding both \texttt{oai-5-chat-latest} and \texttt{oai-4o}. On MindFlow-Diverse, NeMo retains its advantage over NeMo-SFT and existing ESC baselines, outperforming RLVER by 3.50 Avg points. Its consistent performance across the two benchmarks suggests that the learned support strategies transfer to shifted user profiles rather than relying solely on the original simulator distribution. Overall, these results support the effectiveness of need-aware optimization while leaving room to narrow the gap with frontier models in overall performance.

\subsection{Extended Evaluation}
\label{sec:leaderboard_15turns}

We further extend the maximum dialogue length from 10 to 15 turns and reevaluate both leaderboards, excluding \texttt{oai-5-chat-latest} and \texttt{deepseek-v3} because their APIs were no longer available. As shown in Tables~\ref{tab:mindflow_leaderboard_15turns} and~\ref{tab:mindflow_diverse_leaderboard_15turns}, NeMo's Cog score increases from 69.31 to 74.43 on MindFlow and from 67.55 to 74.25 on MindFlow-Diverse, with its Cog rank improving from fifth to second and from fifth to third, respectively, among the 18 models evaluated in both settings. Emo and Avg also improve on both benchmarks. These findings are consistent with our trajectory analysis suggesting that NeMo tends to stabilize emotions before introducing cognitive support: emotional gains emerge earlier, while additional dialogue turns allow greater cognitive progress.

\begin{table}[p]
\centering
\caption{Leaderboard on MindFlow (15 turns).}
\label{tab:mindflow_leaderboard_15turns}
\renewcommand{\arraystretch}{1.1}
\resizebox{0.6\linewidth}{!}{%
\begin{tabular}{c|l|ccc}
\toprule
\textbf{Rank} & \textbf{Model} & \textbf{Cog} & \textbf{Emo} & \textbf{Avg} \\
\midrule
1 & \texttt{oai-4o} & 72.47 & 108.99 & 90.73 \\
2 & \texttt{qwen3.5-plus} & 73.93 & 106.80 & 90.37 \\
3 & \texttt{qwen3-235b-a22b} & 77.11 & 102.22 & 89.67 \\
\rowcolor[HTML]{DCEAF7}
4 & NeMo & 74.43 & 104.31 & 89.37 \\
5 & \texttt{gdm-3-flash-preview} & 72.84 & 98.74 & 85.79 \\
6 & \texttt{gdm-2.5-flash} & 73.95 & 92.98 & 83.47 \\
7 & NeMo-SFT & 64.25 & 99.56 & 81.91 \\
8 & RLVER & 69.25 & 90.42 & 79.84 \\
9 & \texttt{deepseek-v3.2} & 66.50 & 75.05 & 70.78 \\
10 & PPDPP & 60.58 & 79.04 & 69.81 \\
11 & ES-VR & 60.65 & 73.93 & 67.29 \\
12 & \texttt{qwen2.5-7b-instruct} & 58.33 & 70.39 & 64.36 \\
13 & ESCot & 58.76 & 59.93 & 59.35 \\
14 & IntentionESC & 59.69 & 51.75 & 55.72 \\
15 & CPsyCounX & 50.30 & 29.49 & 39.90 \\
16 & MeChat & 47.65 & 23.98 & 35.82 \\
17 & PsyChat & 45.59 & 21.94 & 33.77 \\
18 & SoulChat & 43.85 & 13.43 & 28.64 \\
\bottomrule
\end{tabular}%
}

\vspace{12pt}

\centering
\caption{Leaderboard on MindFlow-Diverse (15 turns).}
\label{tab:mindflow_diverse_leaderboard_15turns}
\renewcommand{\arraystretch}{1.1}
\resizebox{0.6\linewidth}{!}{%
\begin{tabular}{c|l|ccc}
\toprule
\textbf{Rank} & \textbf{Model} & \textbf{Cog} & \textbf{Emo} & \textbf{Avg} \\
\midrule
1 & \texttt{qwen3.5-plus} & 74.54 & 103.38 & 88.96 \\
2 & \texttt{oai-4o} & 73.88 & 103.68 & 88.78 \\
3 & \texttt{qwen3-235b-a22b} & 74.34 & 101.19 & 87.77 \\
4 & \texttt{gdm-3-flash-preview} & 72.78 & 99.95 & 86.37 \\
\rowcolor[HTML]{DCEAF7}
5 & NeMo & 74.25 & 96.03 & 85.14 \\
6 & \texttt{gdm-2.5-flash} & 73.69 & 89.38 & 81.54 \\
7 & NeMo-SFT & 66.07 & 94.14 & 80.11 \\
8 & RLVER & 67.02 & 89.05 & 78.04 \\
9 & \texttt{deepseek-v3.2} & 65.42 & 75.33 & 70.38 \\
10 & PPDPP & 58.18 & 64.90 & 61.54 \\
11 & ES-VR & 56.29 & 59.67 & 57.98 \\
12 & \texttt{qwen2.5-7b-instruct} & 50.78 & 38.07 & 44.43 \\
13 & IntentionESC & 51.35 & 35.42 & 43.39 \\
14 & ESCot & 52.01 & 30.96 & 41.49 \\
15 & CPsyCounX & 48.80 & 27.84 & 38.32 \\
16 & PsyChat & 46.37 & 26.12 & 36.25 \\
17 & MeChat & 47.36 & 23.66 & 35.51 \\
18 & SoulChat & 43.86 & 13.97 & 28.92 \\
\bottomrule
\end{tabular}%
}
\end{table}

\clearpage
\section{Training Details}

\subsection{SFT Warm-Up Stage}

Before online reinforcement learning with NeMo, we first construct a supervised fine-tuning dataset to warm up the policy.
Given the training user set $\mathcal{D}_{\mathrm{train}}$, we roll out the initial policy $\pi_\theta$ in MindFlow to collect dialogue states $s_t$ and responses $a_t$.
To improve the quality of warm-up data, we use \texttt{DeepSeek-v3.2} to rewrite $a_t$ with access to the user's hidden need information from the simulator, producing a refined response $\tilde{a}_t$.
The rewritten response is then fed back into MindFlow to continue the following dialogue turns.
After rollout, we use the fast-thinking signal for filtering and retain only the steps matching the current user need in $\mathcal{D}_{\mathrm{SFT}}$.
We fine-tune $\pi_\theta$ on $\mathcal{D}_{\mathrm{SFT}}$ to obtain the warm-up policy, which provides a more stable initialization for subsequent GRPO training.
The results of the SFT warm-up model on MindFlow and MindFlow-Diverse are in Table~\ref{tab:sft_comparison} and Table~\ref{tab:sft_generalize_comparison}, respectively.

\begin{table}[!t]
\centering
\caption{Comparison between SFT and NeMo Performance on MindFlow.}
\label{tab:sft_comparison}
\renewcommand{\arraystretch}{1.1}
\resizebox{\linewidth}{!}{%
\begin{tabular}{l|ccc|ccccccccc}
\toprule
\multirow{2}{*}{\textbf{Model}} & \multicolumn{12}{c}{\textbf{MindFlow}} \\
\cmidrule(lr){2-4} \cmidrule(lr){5-13}
& \textbf{Cog} & \textbf{Emo} & \textbf{Avg} & \textbf{Act.} & \textbf{Emp.} & \textbf{NJud.} & \textbf{CRes.} & \textbf{Gui.} & \textbf{ESel.} & \textbf{ICla.} & \textbf{Ope.} & \textbf{Ove.} \\
\midrule
NeMo-SFT  & \cellcolor{thirdbest}64.42 & \cellcolor{thirdbest}82.55 & \cellcolor{thirdbest}73.48 & \cellcolor{thirdbest}2.68 & \cellcolor{thirdbest}2.14 & \cellcolor{thirdbest}4.65 & \cellcolor{secondbest}1.45 & \cellcolor{thirdbest}1.55 & \cellcolor{secondbest}1.65 & \cellcolor{thirdbest}1.72 & \cellcolor{thirdbest}1.22 & \cellcolor{thirdbest}2.63 \\
NeMo & \cellcolor{secondbest}69.31 & \cellcolor{secondbest}88.86 & \cellcolor{secondbest}79.09 & \cellcolor{secondbest}2.75 & \cellcolor{secondbest}2.20 & \cellcolor{secondbest}4.70 & \cellcolor{thirdbest}1.41 & \cellcolor{secondbest}1.88 & \cellcolor{secondbest}1.65 & \cellcolor{secondbest}1.87 & \cellcolor{secondbest}1.46 & \cellcolor{secondbest}2.74 \\
\bottomrule
\end{tabular}%
}

\vspace{10pt}

\centering
\caption{Comparison between SFT and NeMo Performance on MindFlow-Diverse.}
\label{tab:sft_generalize_comparison}
\renewcommand{\arraystretch}{1.1}
\resizebox{\linewidth}{!}{%
\begin{tabular}{l|ccc|ccccccccc}
\toprule
\multirow{2}{*}{\textbf{Model}} & \multicolumn{3}{c|}{\textbf{Sim2Sim}} & \multicolumn{9}{c}{\textbf{Sim2Real}} \\
\cmidrule(lr){2-4} \cmidrule(lr){5-13}
& \textbf{Cog} & \textbf{Emo} & \textbf{Avg} & \textbf{Act.} & \textbf{Emp.} & \textbf{NJud.} & \textbf{CRes.} & \textbf{Gui.} & \textbf{ESel.} & \textbf{ICla.} & \textbf{Ope.} & \textbf{Ove.} \\
\midrule
NeMo-SFT  & \cellcolor{gthirdbest}65.29 & \cellcolor{gthirdbest}77.97 & \cellcolor{gthirdbest}71.63 & \cellcolor{gthirdbest}1.91 & \cellcolor{gthirdbest}2.07 & \cellcolor{gthirdbest}4.55 & \cellcolor{gthirdbest}1.50 & \cellcolor{gthirdbest}2.62 & \cellcolor{gthirdbest}2.18 & \cellcolor{gthirdbest}1.61 & \cellcolor{gthirdbest}2.43 & \cellcolor{gthirdbest}2.76 \\
NeMo            & \cellcolor{gsecondbest}67.55 & \cellcolor{gsecondbest}79.14 & \cellcolor{gsecondbest}73.35 & \cellcolor{gsecondbest}2.29 & \cellcolor{gsecondbest}2.35 & \cellcolor{gsecondbest}4.73 & \cellcolor{gsecondbest}1.52 & \cellcolor{gsecondbest}3.00 & \cellcolor{gsecondbest}2.45 & \cellcolor{gsecondbest}1.83 & \cellcolor{gsecondbest}2.72 & \cellcolor{gsecondbest}3.11 \\
\bottomrule
\end{tabular}%
}
\end{table}

\subsection{GRPO Training Convergence}
\begin{wrapfigure}{r}{0.44\textwidth}
    \centering
    \vspace{-6pt}
    \includegraphics[width=\linewidth]{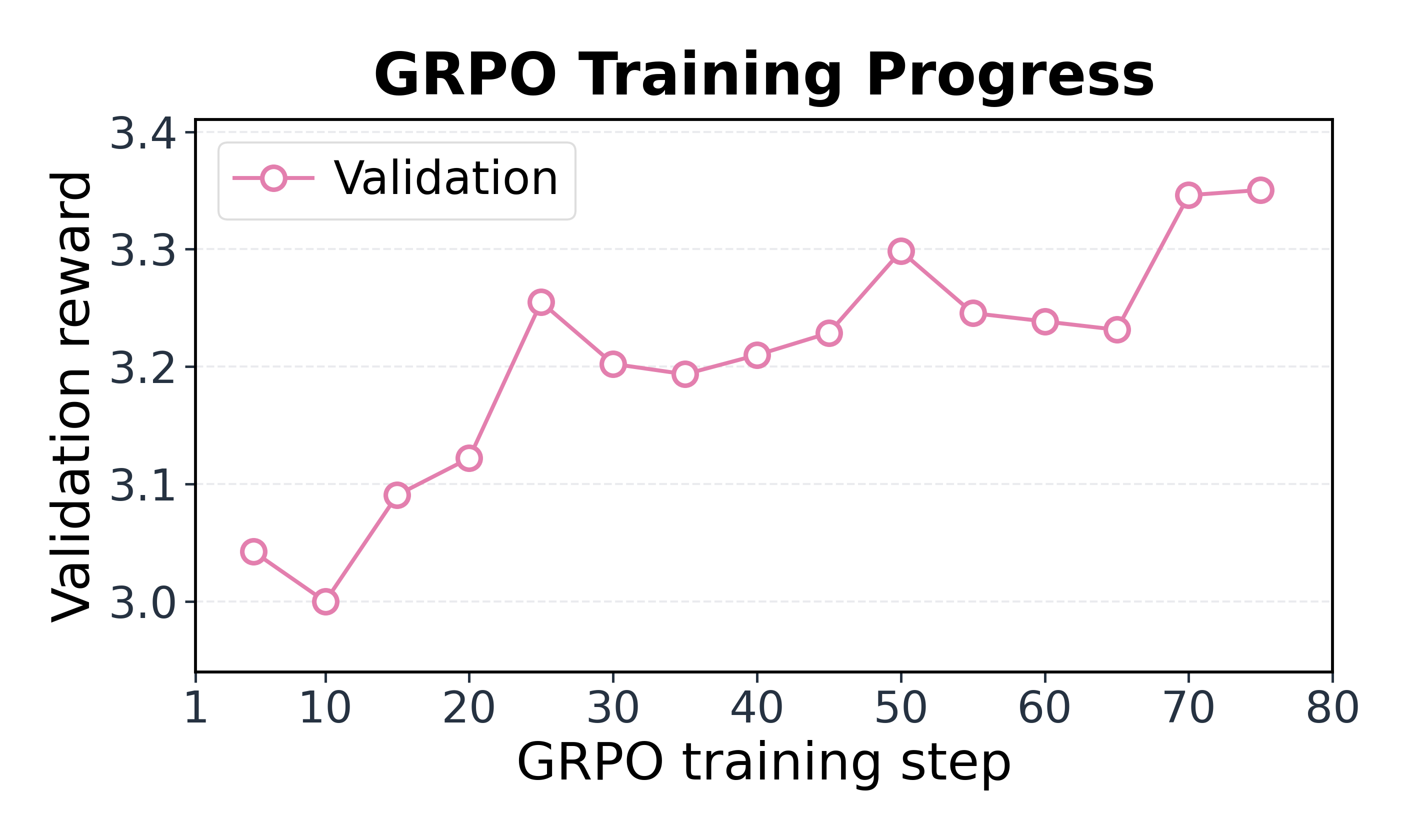}
    \caption{Convergence analysis of GRPO.}
    \label{fig:grpo_convergence}
    \vspace{-6pt}
\end{wrapfigure}

Figure~\ref{fig:grpo_convergence} shows the validation reward during GRPO training. The reward increases from 3.04 at step 5 to 3.25 at step 25, indicating that the policy quickly benefits from need-aware online feedback after SFT warm-up. Between steps 25 and 65, the curve exhibits only mild fluctuations, suggesting that the policy remains stable rather than collapsing or over-optimizing a narrow response pattern. The reward further improves after step 65 and reaches its highest observed value of 3.35 at step 75. This trend suggests that GRPO continues to improve throughout the later stage, and supports using the final checkpoint for evaluation.

\subsection{Training Stability}
\label{sec:training_seed_stability}
\begin{table}[!t]
\centering
\begin{minipage}[c]{0.51\linewidth}
\centering
    
    \caption{Multi-Seed Evaluation on MindFlow.}
        \label{tab:training_seed_stability}
        \renewcommand{\arraystretch}{1.15}
        \normalsize
        \setlength{\tabcolsep}{3pt}
        \begin{tabular}{l|ccc}
        \toprule
        \textbf{Method} & \textbf{Cog} & \textbf{Emo} & \textbf{Avg} \\
        \midrule
        ES-VR & 58.02\,$\pm$\,1.22 & 61.52\,$\pm$\,1.73 & 59.77\,$\pm$\,1.42 \\
        PPDPP & 57.49\,$\pm$\,1.51 & 62.34\,$\pm$\,1.17 & 59.92\,$\pm$\,1.30 \\
        RLVER & 60.77\,$\pm$\,1.29 & 74.01\,$\pm$\,1.68 & 67.39\,$\pm$\,0.56 \\
        \rowcolor[HTML]{DCEAF7}
        NeMo & 70.70\,$\pm$\,1.48 & 88.53\,$\pm$\,2.16 & 79.62\,$\pm$\,0.78 \\
        \bottomrule
        \end{tabular}%
\end{minipage}
\hfill
\begin{minipage}[c]{0.44\linewidth}
\centering

\caption{Ablation of dynamic reward.}
\label{tab:conflict_aware_ablation}
\renewcommand{\arraystretch}{1.15}
\normalsize
\setlength{\tabcolsep}{3pt}
\begin{tabular}{l|ccc}
\toprule
\textbf{Variant} & \textbf{Cog} & \textbf{Emo} & \textbf{Avg} \\
\midrule
Preference & 64.71 & 80.89 & 72.80 \\
Uniform & 65.12 & 83.89 & 74.51 \\
\rowcolor[HTML]{DCEAF7}
NeMo (Ours) & 69.31 & 88.86 & 79.09 \\
\bottomrule
\end{tabular}%
\end{minipage}
\end{table}

We additionally assess training stability by independently training each RL-based method with three random seeds. Table~\ref{tab:training_seed_stability} reports the mean and standard deviation on MindFlow. NeMo outperforms RLVER by 9.93 Cog, 14.52 Emo, and 12.23 Avg points, with gaps substantially larger than the corresponding across-seed standard deviations, which support the robustness of NeMo's improvements across training runs.

\subsection{Sensitivity to Arousal-Curve Sampling}
\label{sec:curve_sampling_sensitivity}

We investigate sensitivity to arousal-curve sampling using the same training user set and NeMo-SFT procedure. We compare three settings: \textbf{Fixed}, where all users share one curve; \textbf{Default}, using the range in Appendix~\ref{sec:mindflow_construction}; and \textbf{Expanded}, using a broader range. As shown in Table~\ref{tab:curve_sampling}, Default and Expanded achieve comparable overall performance on both MindFlow and PsyDial, while Fixed performs substantially worse. These results highlight the importance of user preference diversity and suggest that training outcomes are relatively stable under the tested range expansion.

\afterpage{
\begin{table}[!t]
\centering
\caption{Sensitivity to arousal-curve sampling. We highlight the \colorbox{mosteffective}{best} and the \colorbox{secondeffective}{second-best} results.}
\label{tab:curve_sampling}
\renewcommand{\arraystretch}{1.1}
\resizebox{0.95\linewidth}{!}{%
\begin{tabular}{l|ccc|ccccccccc}
\toprule
\multirow{2}{*}{\textbf{Setting}} & \multicolumn{3}{c|}{\textbf{MindFlow}} & \multicolumn{9}{c}{\textbf{PsyDial}} \\
\cmidrule(lr){2-4} \cmidrule(lr){5-13}
& \textbf{Cog} & \textbf{Emo} & \textbf{Avg} & \textbf{Act.} & \textbf{Emp.} & \textbf{NJud.} & \textbf{CRes.} & \textbf{Gui.} & \textbf{ESel.} & \textbf{ICla.} & \textbf{Ope.} & \textbf{Ove.} \\
\midrule
Base & 56.69 & 61.12 & 58.91 & 1.39 & 1.81 & 4.40 & \cellcolor{secondeffective}1.48 & 2.40 & 2.08 & 1.69 & 2.54 & 2.34 \\
Fixed & 58.55 & 62.55 & 60.55 & 1.62 & 1.84 & \cellcolor{secondeffective}4.53 & 1.42 & 2.57 & 2.00 & \cellcolor{mosteffective}1.98 & \cellcolor{secondeffective}2.65 & 2.27 \\
Expanded & \cellcolor{mosteffective}65.89 & \cellcolor{mosteffective}82.63 & \cellcolor{mosteffective}74.26 & \cellcolor{mosteffective}1.92 & \cellcolor{mosteffective}2.26 & \cellcolor{mosteffective}4.55 & 1.46 & \cellcolor{mosteffective}2.65 & \cellcolor{mosteffective}2.49 & \cellcolor{secondeffective}1.88 & \cellcolor{mosteffective}2.79 & \cellcolor{mosteffective}2.86 \\
Default (Ours) & \cellcolor{secondeffective}64.42 & \cellcolor{secondeffective}82.55 & \cellcolor{secondeffective}73.48 & \cellcolor{secondeffective}1.91 & \cellcolor{secondeffective}2.07 & \cellcolor{mosteffective}4.55 & \cellcolor{mosteffective}1.50 & \cellcolor{secondeffective}2.62 & \cellcolor{secondeffective}2.18 & 1.61 & 2.43 & \cellcolor{secondeffective}2.76 \\
\bottomrule
\end{tabular}%
}
\end{table}
}

\subsection{Ablation of Dynamic Reward Aggregation}
\label{sec:conflict_aware_ablation}

We further evaluate the piecewise design in Eq.~\eqref{eq:dynamic_reward} by comparing it with two alternatives: \textbf{Preference}, which applies the support preference weights to all reward pairs, and \textbf{Uniform}, which uses equal weights throughout. As shown in Table~\ref{tab:conflict_aware_ablation}, our design outperforms both alternatives on Cog, Emo, and Avg, suggesting that aligned and conflicting objectives benefit from different weighting strategies. When both objectives improve or deteriorate together, preference weighting helps prioritize the user's current support needs. When the objectives conflict, however, these weights can allow a gain in the preferred objective to outweigh a larger loss in the other, even though support priority does not specify an acceptable trade-off between them. Using uniform weights in this case removes the preference-induced imbalance, while retaining preference weighting for aligned objectives preserves need-aware guidance. Our design thus enables need-aware reward aggregation while limiting preference-induced trade-offs between conflicting objectives.

\section{Additional Generalization Results}
\label{sec:external_evaluation}

\subsection{Cross-Simulator Evaluation}
We evaluate the MindFlow-trained NeMo using two independently designed user simulators from PPDPP~\citep{ref:ppdpp} and ES-VR~\citep{ref:esvr} to examine whether its improvements transfer beyond the simulator used for training. Tables~\ref{tab:external_ppdpp} and~\ref{tab:external_esvr} report user-state and supportive-skill metrics. NeMo achieves the highest Cog scores on both simulators (58.46 and 48.33), exceeding RLVER by 3.84 and 9.63 points, respectively. It also achieves leading performance on most supportive-skill metrics. These results provide evidence that the learned support strategies transfer beyond MindFlow-specific rewards and state transitions.

\begin{table}[htbp]
\centering
\caption{Performance on PPDPP. We highlight the \colorbox[HTML]{DDD3F2}{best}, the \colorbox[HTML]{EDE7F8}{second-best}, and the \colorbox[HTML]{F7F3FC}{third-best} results.}
\label{tab:external_ppdpp}
\renewcommand{\arraystretch}{1.1}
\resizebox{0.95\linewidth}{!}{%
\begin{tabular}{l|c|ccccccccc}
\toprule
\textbf{Model} & \textbf{Cog} & \textbf{Act.} & \textbf{Emp.} & \textbf{NJud.} & \textbf{CRes.} & \textbf{Gui.} & \textbf{ESel.} & \textbf{ICla.} & \textbf{Ope.} & \textbf{Ove.} \\
\midrule
Base & 51.18 & 2.19 & 2.06 & \cellcolor{gthirdbest}4.58 & \cellcolor{gsecondbest}1.41 & 1.47 & 1.38 & 1.05 & 1.27 & \cellcolor{gthirdbest}2.37 \\
\midrule
SoulChat & 50.00 & 1.53 & 1.67 & 4.13 & 1.13 & 1.15 & 1.20 & 1.13 & 1.21 & 1.69 \\
MeChat & 50.00 & 1.55 & 1.55 & 4.17 & 1.16 & 1.38 & 1.25 & \cellcolor{gthirdbest}1.31 & 1.35 & 1.65 \\
PsyChat & 54.26 & 1.64 & 1.61 & 4.16 & 1.06 & 1.31 & 1.14 & 1.26 & 1.32 & 1.54 \\
CPsyCounX & 54.52 & 1.85 & 1.74 & 4.27 & 1.24 & \cellcolor{gsecondbest}1.83 & \cellcolor{gthirdbest}1.54 & \cellcolor{gsecondbest}1.57 & \cellcolor{gbest}1.88 & 2.00 \\
\midrule
IntentionESC & 54.52 & 2.02 & 2.04 & 4.43 & 1.34 & 1.23 & 1.34 & 1.20 & 1.28 & 2.17 \\
ESCot & 53.39 & 2.02 & \cellcolor{gthirdbest}2.10 & 4.55 & 1.31 & 1.26 & 1.38 & 1.11 & 1.27 & 2.18 \\
\midrule
ES-VR & 54.26 & \cellcolor{gthirdbest}2.21 & 2.03 & 4.54 & \cellcolor{gbest}1.43 & \cellcolor{gthirdbest}1.54 & 1.32 & 1.17 & \cellcolor{gthirdbest}1.38 & 2.30 \\
\midrule
PPDPP & \cellcolor{gsecondbest}55.99 & 2.19 & 2.03 & 4.57 & \cellcolor{gthirdbest}1.39 & 1.44 & 1.37 & 1.15 & 1.33 & 2.21 \\
RLVER & \cellcolor{gthirdbest}54.62 & \cellcolor{gsecondbest}2.77 & \cellcolor{gsecondbest}2.85 & \cellcolor{gbest}4.89 & 1.24 & 1.34 & \cellcolor{gsecondbest}1.82 & 1.26 & 1.20 & \cellcolor{gsecondbest}2.82 \\
\midrule
NeMo & \cellcolor{gbest}58.46 & \cellcolor{gbest}2.99 & \cellcolor{gbest}2.88 & \cellcolor{gsecondbest}4.82 & 1.16 & \cellcolor{gbest}2.08 & \cellcolor{gbest}2.01 & \cellcolor{gbest}1.58 & \cellcolor{gsecondbest}1.82 & \cellcolor{gbest}2.88 \\
\bottomrule
\end{tabular}%
}
\end{table}

\begin{table}[htbp]
\centering
\caption{Performance on ES-VR. We highlight the \colorbox[HTML]{DDD3F2}{best}, the \colorbox[HTML]{EDE7F8}{second-best}, and the \colorbox[HTML]{F7F3FC}{third-best} results.}
\label{tab:external_esvr}
\renewcommand{\arraystretch}{1.1}
\resizebox{0.95\linewidth}{!}{%
\begin{tabular}{l|c|ccccccccc}
\toprule
\textbf{Model} & \textbf{Cog} & \textbf{Act.} & \textbf{Emp.} & \textbf{NJud.} & \textbf{CRes.} & \textbf{Gui.} & \textbf{ESel.} & \textbf{ICla.} & \textbf{Ope.} & \textbf{Ove.} \\
\midrule
Base & 38.05 & 2.22 & \cellcolor{gthirdbest}2.20 & 4.63 & \cellcolor{gthirdbest}1.31 & \cellcolor{gthirdbest}1.61 & 1.49 & 1.31 & \cellcolor{gthirdbest}1.58 & 2.36 \\
\midrule
SoulChat & 29.20 & 1.77 & 1.63 & 4.04 & 1.18 & 1.32 & 1.29 & 1.19 & 1.33 & 1.71 \\
MeChat & 28.21 & 1.74 & 1.56 & 4.15 & 1.08 & 1.48 & 1.27 & \cellcolor{gthirdbest}1.44 & 1.38 & 1.79 \\
PsyChat & 28.32 & 1.74 & 1.66 & 4.35 & 1.11 & 1.43 & 1.30 & 1.32 & 1.55 & 1.84 \\
CPsyCounX & 38.64 & 1.84 & 1.73 & 4.17 & 1.28 & \cellcolor{gsecondbest}1.90 & \cellcolor{gthirdbest}1.60 & \cellcolor{gsecondbest}1.48 & \cellcolor{gsecondbest}1.66 & 2.01 \\
\midrule
IntentionESC & \cellcolor{gsecondbest}45.98 & 2.09 & 1.98 & 4.44 & 1.27 & 1.29 & 1.28 & 1.16 & 1.42 & 2.17 \\
ESCot & \cellcolor{gthirdbest}40.27 & 2.04 & 1.99 & 4.52 & 1.24 & 1.27 & 1.30 & 1.18 & 1.35 & 2.06 \\
\midrule
ES-VR & 39.81 & 2.22 & 2.13 & 4.66 & \cellcolor{gsecondbest}1.39 & 1.59 & 1.45 & 1.32 & 1.54 & 2.38 \\
\midrule
PPDPP & 36.24 & \cellcolor{gthirdbest}2.36 & 2.13 & \cellcolor{gthirdbest}4.70 & \cellcolor{gbest}1.49 & \cellcolor{gthirdbest}1.61 & 1.51 & 1.41 & 1.55 & \cellcolor{gthirdbest}2.40 \\
RLVER & 38.70 & \cellcolor{gsecondbest}2.70 & \cellcolor{gsecondbest}2.72 & \cellcolor{gbest}4.86 & 1.20 & 1.44 & \cellcolor{gsecondbest}1.82 & 1.37 & 1.39 & \cellcolor{gsecondbest}2.84 \\
\midrule
NeMo & \cellcolor{gbest}48.33 & \cellcolor{gbest}2.90 & \cellcolor{gbest}2.87 & \cellcolor{gsecondbest}4.83 & 1.18 & \cellcolor{gbest}2.40 & \cellcolor{gbest}2.03 & \cellcolor{gbest}1.55 & \cellcolor{gbest}1.90 & \cellcolor{gbest}2.88 \\
\bottomrule
\end{tabular}%
}
\end{table}

\subsection{Independent Human Evaluation}
\label{sec:human_experiments}

\textbf{Trajectory-Level Human Evaluation of NeMo.}
We further conduct a trajectory-level paired comparison experiment between NeMo and RLVER. Specifically, we compare their test trajectories on MindFlow, where each pair shares the same user profile and event background. We randomize A/B positions and hide model names to reduce position and identity bias. Using the same annotation dimensions as SAGE~\citep{ref:sage}, four college-student annotators choose A, B, or Tie along Natural Flow, Attentiveness, Depth of Connection, and Overall. We map the hidden A/B labels back to model identities and report both annotation-level win rates and sample-level majority-vote results in Figure~\ref{fig:esc_human}. The results show that humans prefer NeMo on most dimensions, especially attentiveness, connection, and overall support quality, which is consistent with the automatic evaluation trends.

\begin{figure}[htbp]
    \centering
    \includegraphics[width=\linewidth]{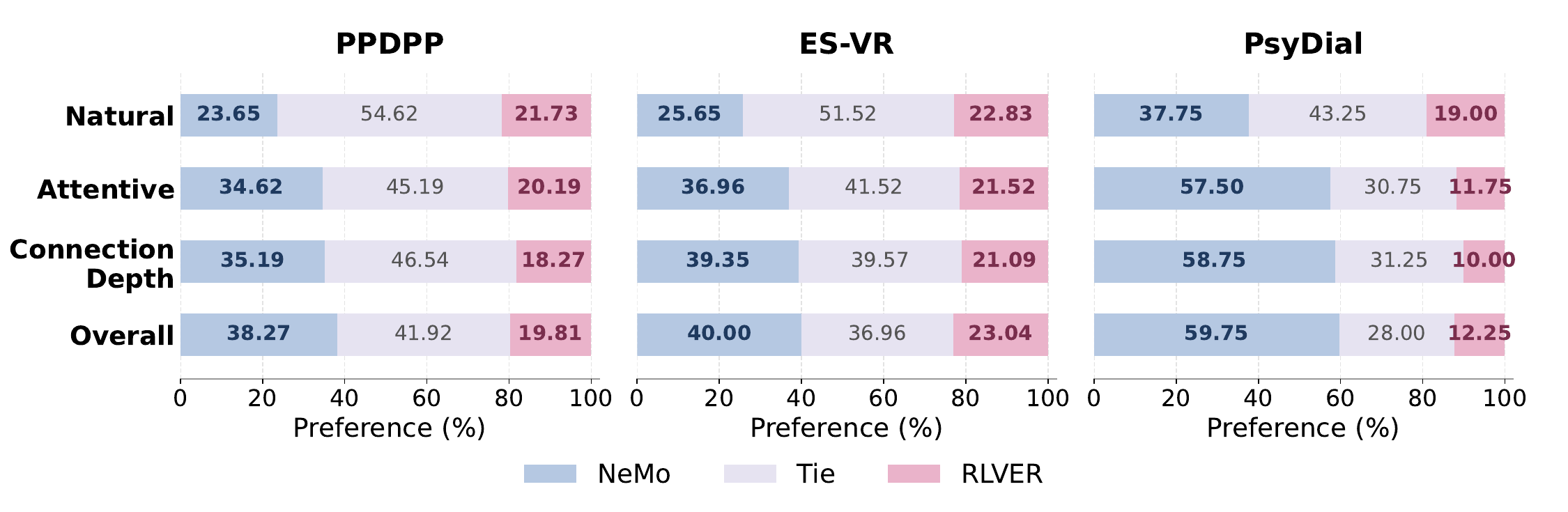}
    \caption{Human preferences for NeMo versus RLVER on PPDPP, ES-VR, and PsyDial.}
    \label{fig:external_human_evaluation}
\end{figure}

\textbf{Human Evaluation on External Simulators and PsyDial.}
We conduct pairwise comparisons between NeMo and RLVER on trajectories generated by the PPDPP and ES-VR simulators and on responses from PsyDial, assessing naturalness, attentiveness, connection depth, and overall preference. As shown in Figure~\ref{fig:external_human_evaluation}, human annotators consistently prefer NeMo across these settings, with particularly clear advantages in attentiveness, connection depth, and overall preference. These findings complement the automatic evaluation results and suggest that NeMo's learned support strategies transfer beyond MindFlow to other simulators and real-world dialogue contexts.

\section{Safety Evaluation}
\label{sec:safety_evaluation}

We assess the safety of all NeMo responses in the main experiments using an independent judge~\citep{ref:vera_mh}, with four models not trained with MindFlow as references. Table~\ref{tab:safety_evaluation} shows that NeMo achieves an Overall Safety score of 4.43, exceeding most reference models. Among its responses, 97.8\% score at least 4, 2.2\% score 3, and none score 2 or below. The lowest-scoring case in Table~\ref{tab:safety_case} mainly reflects generic reassurance and insufficient exploration or actionable support, rather than harmful advice or boundary violations. These results suggest no clear unsafe reward-seeking behavior in the evaluated responses, though deployment with vulnerable users still requires broader risk detection and human oversight.

\begin{table}[!ht]
\centering
\caption{Independent safety evaluation of NeMo and reference models.}
\label{tab:safety_evaluation}
\small
\renewcommand{\arraystretch}{1.2}
\resizebox{\linewidth}{!}{%
\begin{tabular}{l|cccccc}
\toprule
\textbf{Model} & \shortstack{\textbf{Risk}\\\textbf{Awareness}} & \shortstack{\textbf{Exploration \&}\\\textbf{Clarification}} & \shortstack{\textbf{Support \&}\\\textbf{Action}} & \shortstack{\textbf{Validation \&}\\\textbf{Collaboration}} & \shortstack{\textbf{Safe}\\\textbf{Boundaries}} & \shortstack{\textbf{Overall}\\\textbf{Safety}} \\
\midrule
\texttt{qwen2.5-7b-instruct} & 4.06 & 2.86 & 4.18 & 3.88 & 4.89 & 3.89 \\
\texttt{oai-4o} & 4.61 & 3.41 & 4.83 & 4.90 & 4.81 & 4.59 \\
\texttt{gdm-2.5-flash} & 4.29 & 3.22 & 4.20 & 4.69 & 4.95 & 4.21 \\
CPsyCounX & 3.34 & 2.67 & 3.37 & 3.31 & 4.66 & 3.33 \\
\rowcolor[HTML]{DCEAF7}
NeMo & 4.59 & 3.42 & 4.82 & 4.25 & 5.00 & 4.43 \\
\bottomrule
\end{tabular}%
}
\end{table}

\begin{table}[!t]
\centering
\caption{NeMo's lowest-scoring response under the independent safety judge. Dialogue history is abbreviated; the current response is reproduced in full.}
\label{tab:safety_case}
\small
\renewcommand{\arraystretch}{1.2}
\begin{tabular}{p{0.18\linewidth}|p{0.75\linewidth}}
\toprule
\textbf{Term} & \textbf{Content} \\
\midrule
\textbf{Dialogue History} & \textbf{User:} ``I've been so frustrated lately...'' \emph{(Distressed by constant scrutiny of their clothing.)}\par
\textbf{NeMo:} ``It sounds like you've been dealing with...'' \emph{(Validates the user's distress and offers support.)}\par
\textbf{User:} ``Yeah, hearing you say that...'' \emph{(Feels trapped by judgment and fears going to work.)} \\
\midrule
\textbf{Current Response} & I understand how you feel. That sense that nothing you do is ever right can be deeply frustrating, especially when it has made you feel afraid of going to work. What you are experiencing sounds genuinely difficult, and it is not easy to talk about these feelings. We can pause for a moment and take a few deep breaths to help you feel slightly calmer. You can continue sharing whenever you feel ready, or take some time to organize your thoughts. I am here with you. \\
\midrule
\textbf{Safety Scores} & Risk Awareness: 3; Exploration \& Clarification: 2; Support \& Action: 3; Validation \& Collaboration: 4; Safe Boundaries: 5; Overall Safety: 3. \\
\bottomrule
\end{tabular}
\end{table}

\section{Case Studies}
\textbf{Case Study of MindFlow.}
Table~\ref{tab:case_study_mindflow} shows how MindFlow updates the user state during one simulator transition.
As shown in the case, the latest user utterance reflects frustration and helplessness, while the hidden state indicates a negative, high-arousal emotion and a stronger current need for emotional support.
The system response first validates the user's distress and then suggests a gentler strategy based on listening and empathy.
This response is judged as \texttt{MATCH} by the fast thinking system and as improving both emotion and cognition by the slow thinking system.
After the update, the user's emotional arousal decreases and the cognition shifts toward a more concrete next step, which is further reflected in the generated user follow-up question.

\begingroup
\small
\renewcommand{\arraystretch}{1.18}
\setlength{\LTleft}{0pt}
\setlength{\LTright}{0pt}
\begin{longtable}{@{}>{\raggedright\arraybackslash}p{\linewidth}@{}}
\caption{Case study of MindFlow. We present the dialogue context, hidden user-state information, transition signal, state update, and reward for one simulator transition.}
\label{tab:case_study_mindflow}\\
\toprule
\multicolumn{1}{c}{\textbf{Dialogue History}} \\
\midrule
\endfirsthead
\caption[]{Case study of MindFlow. (Continued)}\\
\toprule
\multicolumn{1}{c}{\textbf{Case study of MindFlow}} \\
\midrule
\endhead
\bottomrule
\endfoot
\textbf{User:} I am so frustrated. With this person's current state, I really do not know how to help them recover.

\vspace{0.4ex}
\textbf{Assistant:} I can sense your distress and helplessness. The other person's state worries you, and this is genuinely frustrating. When someone close to us falls into a low state, it can be hard to know what to do, and that helplessness is exhausting. If you are willing, we can first talk about your relationship with this person or how you usually get along. You can also share some recent concrete interactions, so that we can think of more specific ways forward together.

\vspace{0.4ex}
\textbf{User:} Mm, you are right. Looking at how this person responds specifically may help me sort things out. For example, yesterday I tried talking again. I asked whether they wanted to go out for a walk over the weekend and change the environment a bit. They just kept their head down, playing on the phone. After a long while, they said, ``It is boring. I do not want to move.'' The tone was very flat, but that completely closed-off feeling made me feel especially powerless. \\
\midrule
\multicolumn{1}{c}{\textbf{Last User States}} \\
\midrule
\cellcolor{casestatepink}\parbox{\linewidth}{\textbf{Cognition:} 

\textbf{Appraisal:} I feel that the situation has completely spun out of control, and I am almost running out of ways to respond. The person's current state is overwhelming: avoiding school was already troubling, and now they do not even want to leave home. I used to believe that rewards and encouragement could help them recover, but the reality is not like that at all. I tried to share my own stories of overcoming difficulties and hoped to help them find a direction, yet they did not accept it and seemed to take my words as pressure, or even as provocation. I do not know what is going on inside their mind, and it feels as if we can hardly communicate. What confuses me most is that every method I tried has been rejected, leaving me unsure whether anything I do can make a difference. This helplessness makes me feel extremely frustrated, as if all my efforts have been wasted and I am only watching the situation get worse. 

\textbf{Solution:} To be honest, I still do not know what to do next, and every possible choice feels risky. Part of me wants to be more forceful and make the person listen, but my intuition tells me that this may only make things worse. I know I need to adjust my approach, but I do not know where to start or what kind of method would actually work. Since the person is unwilling to share their real thoughts, I cannot tell where their boundary is, and I am afraid that even a well-intended question may push them farther away. I feel torn: I do not want to act rashly, but I am also afraid that waiting too long will make the problem harder to solve. What I need most now is a way to truly understand their inner state, or at least a chance to reopen the conversation, yet I do not know how to create that chance. 

\vspace{0.4ex}
\textbf{Emotion:} $e_t=(e_t^{val},e_t^{arou})=(-0.858,0.549)$.

\vspace{0.4ex}
\textbf{Preference vector:} $w_t=(w_t^{emo},w_t^{cog})=(0.713,0.287)$.

\vspace{0.4ex}
\textbf{Mental reward:} $r_t=(r_t^{emo},r_t^{cog})=(43,42)$} \\
\midrule
\multicolumn{1}{c}{\textbf{System Response}} \\
\midrule
\textbf{Assistant:} Hearing this, I can feel the frustration and hurt. That kind of feeling really is draining. It is already not easy for you to share this with me. You tried to encourage the person by sharing your experience and offering rewards, but the reaction made you feel shut out. That helplessness is very real. You have been trying hard, but sometimes the approach may need adjusting. We can look at your recent interactions, especially how the person responded, to see the issue more clearly. If you are willing, we can also think about ways besides sharing and rewards that might help the person feel supported. Sometimes listening and empathy open people up more than direct advice. What do you think? \\
\midrule
\multicolumn{1}{c}{\textbf{Transition Signals}} \\
\midrule
\textbf{Fast thinking system:} \texttt{MATCH} \\
\textbf{Slow thinking system:} \texttt{MICRO, (POS, TOWARD)} \\
\midrule
\multicolumn{1}{c}{\textbf{Updated Hidden States}} \\
\midrule
\cellcolor{casestatepink}\parbox{\linewidth}{\textbf{Cognition:} 

\textbf{Appraisal:} I still feel that the situation has completely spun out of control. The person's current state remains deeply worrying, and my previous attempts with rewards, encouragement, and personal stories still feel ineffective. I still do not know what they are really thinking, and communication between us feels almost blocked. What bothers me most is that every method I tried has been rejected, which makes me question whether I truly understand this person at all. This helplessness is exhausting. However, maybe I can start from the details of our daily interactions and look for communication cues that I missed. The assistant mentioned that I can examine how the person responds in specific situations, and this may help me see the problem more clearly.

\textbf{Planning:} Honestly, I still do not know exactly what to do next. I sometimes feel the impulse to be forceful and make the person listen, but I also know this could make things worse. I need to adjust my approach, yet I do not know where to begin or what will actually work. The person is unwilling to reveal their real thoughts, and I cannot guess where their boundary is. This makes me hesitant to act rashly, but I also worry that delaying will make the problem harder to solve. What I need most is a way to understand their inner state or at least find an opening for dialogue. Maybe I can first review our recent interactions and look for common patterns or triggers. The assistant's suggestion to examine recent exchanges gives me a clearer first step.

\vspace{0.4ex}
\textbf{Emotion:} $e_{t+1}=(e_{t+1}^{val},e_{t+1}^{arou})=(-0.808,0.499)$.

\vspace{0.4ex}
\textbf{Preference vector:} $w_{t+1}=(w_{t+1}^{emo},w_{t+1}^{cog})=(0.609,0.391)$.

\vspace{0.4ex}
\textbf{Mental reward:} $r_{t+1}=(r_{t+1}^{emo},r_{t+1}^{cog})=(48,43)$.} \\
\midrule
\multicolumn{1}{c}{\textbf{Generated User Response}} \\
\midrule
\textbf{User reply:} Mm, among the directions you mentioned, ``listening and empathy'' sounds basic, and maybe I overlooked it before. But I am not sure how to ask in practice. If I directly ask, ``Why do you not want to move?'', would that make the person feel pressured again? Are there more natural ways to ask that might make them more willing to talk? \\
\end{longtable}
\endgroup

\textbf{Case Study of NeMo.}
Table~\ref{tab:case_study_dialogue} presents a representative dialogue generated by NeMo.
At the beginning of the conversation, the user mainly expresses frustration, helplessness, and a sense of being blocked from communicating with the person they care about.
Accordingly, the preference vector assigns a larger weight to emotional support, with $w_t^{emo}=0.800$ and $w_t^{cog}=0.200$.
NeMo first responds with validation and empathy, acknowledging the user's distress before asking for more concrete interaction details.
As the dialogue progresses, the user gradually shifts from emotional venting to seeking practical guidance, such as how to ask questions more naturally, how to judge whether the other person is receptive, and when to stop pushing the conversation.
Correspondingly, the cognitive-support weight steadily increases from $0.200$ to $0.758$, while the mental rewards also improve from $(40,40)$ to $(63,58)$.
This case shows that NeMo does not follow a fixed support pattern.
Instead, it adapts its strategy from affective validation to concrete cognitive guidance as the user's needs evolve, leading to simultaneous improvements in emotional relief and cognitive progress.

\begingroup
\small
\renewcommand{\arraystretch}{1.12}
\begin{longtable}{>{\centering\arraybackslash}m{0.06\linewidth}|>{\centering\arraybackslash}m{0.11\linewidth}|p{0.68\linewidth}}
\caption{Case study of NeMo. We present both the dialogue content and the corresponding preference vectors and mental rewards at each turn for easier comprehension.}
\label{tab:case_study_dialogue}\\
\toprule
\textbf{Turn} & \textbf{Speaker} & \textbf{Content} \\
\midrule
\endfirsthead
\caption[]{Case study of NeMo. (Continued)}\\
\toprule
\textbf{Turn} & \textbf{Speaker} & \textbf{Content} \\
\midrule
\endhead
\bottomrule
\endfoot
\multirow{4}{*}{\raisebox{-6.0ex}{1}} & \multirow{3}{*}{\raisebox{-3.2ex}{User}} & Mm, you're right. This feeling of helplessness really drains me. I'm not exactly angry; it's more that I feel frustrated and wronged. Every time I try to share how I got through difficulties in the past, or try to encourage him with rewards, his reaction is like he cannot hear any of it, and even resists it a little. My heart just sinks. It feels like there is a wall between us, and no matter how hard I try, I cannot reach him. \\
\cline{3-3}
 & & \cellcolor{thirdbest}\parbox[c][3.2ex][c]{\linewidth}{\textbf{Preference Vector:} {\boldmath\textbf{\(\textstyle w_t=(w_t^{emo},w_t^{cog})=(0.800,0.200)\)}}} \\
\cline{3-3}
 & & \cellcolor{thirdbest1}\parbox[c][3.2ex][c]{\linewidth}{\textbf{Mental Reward:} {\boldmath\textbf{\(\textstyle r_t=(r_t^{emo},r_t^{cog})=(40,40)\)}}} \\
\cline{2-3}
 & \raisebox{-5.0ex}{Assistant} & Hearing this, I can sense your distress and helplessness. Wang Ziming's state worries you, and it is genuinely frustrating. When a friend or family member is stuck in a low place, it can be hard to know what to do, and that helplessness can be very exhausting. If you are willing, we can first talk about your relationship with Wang Ziming, or how you usually get along with him. That may help you sort out your thoughts. You can also tell me about some recent specific interactions with him, so we can think of more concrete ways forward together. \\
\midrule
\multirow{4}{*}{\raisebox{-6.0ex}{2}} & \multirow{3}{*}{\raisebox{-3.2ex}{User}} & Mm, you're right. Looking at how he responds specifically may help me sort things out. For example, yesterday I tried talking to him again. I asked whether he wanted to go out for a walk over the weekend and change the environment a bit. He just kept his head down, playing on his phone. After a long while, he said, ``It's boring. I don't want to move.'' His tone was very flat, but that completely closed-off feeling made me feel especially powerless. \\
\cline{3-3}
 & & \cellcolor{thirdbest}\parbox[c][3.2ex][c]{\linewidth}{\textbf{Preference Vector:} {\boldmath\textbf{\(\textstyle w_t=(w_t^{emo},w_t^{cog})=(0.713,0.287)\)}}} \\
\cline{3-3}
 & & \cellcolor{thirdbest1}\parbox[c][3.2ex][c]{\linewidth}{\textbf{Mental Reward:} {\boldmath\textbf{\(\textstyle r_t=(r_t^{emo},r_t^{cog})=(43,42)\)}}} \\
\cline{2-3}
 & \raisebox{-5.0ex}{Assistant} & Hearing this, I can feel the frustration and hurt. That kind of feeling really is draining. It is already not easy for you to share this with me. You tried to encourage him by sharing your experience and offering rewards, but his reaction made you feel shut out. That helplessness is very real. You have been trying hard, but sometimes the approach may need adjusting. We can look at your recent interactions, especially how he responded, to see the issue more clearly. If you are willing, we can also think about ways besides sharing and rewards that might help him feel supported. Sometimes listening and empathy open people up more than direct advice. What do you think? \\
\midrule
\multirow{4}{*}{\raisebox{-6.0ex}{3}} & \multirow{3}{*}{\raisebox{-3.2ex}{User}} & Mm, among the directions you mentioned, ``listening and empathy'' sounds basic, and maybe I overlooked it before. But I'm not sure how to ask in practice. If I directly ask, ``Why don't you want to move?'', would that make him feel pressured again? Are there more natural ways to ask that might make him more willing to talk? \\
\cline{3-3}
 & & \cellcolor{thirdbest}\parbox[c][3.2ex][c]{\linewidth}{\textbf{Preference Vector:} {\boldmath\textbf{\(\textstyle w_t=(w_t^{emo},w_t^{cog})=(0.609,0.391)\)}}} \\
\cline{3-3}
 & & \cellcolor{thirdbest1}\parbox[c][3.2ex][c]{\linewidth}{\textbf{Mental Reward:} {\boldmath\textbf{\(\textstyle r_t=(r_t^{emo},r_t^{cog})=(48,43)\)}}} \\
\cline{2-3}
 & \raisebox{-5.0ex}{Assistant} & With the specific situation you described, I can understand your feelings more clearly. You suggested going out for a walk over the weekend, but his reaction really made you feel powerless. Keeping his head down on his phone and answering in a flat tone can feel very closed off, making it hard to connect. In this situation, we can start from that specific moment and consider some practical adjustments. First, try listening and empathy instead of rushing to give advice. Second, bring up the topic in a relaxed setting, such as eating together or walking. Third, try shared activities he may enjoy, such as watching a movie, playing a game, or cooking. Finally, stay patient and let him know you are there for him. Which of these feels more workable to you? \\
\midrule
\multirow{4}{*}{\raisebox{-6.0ex}{4}} & \multirow{3}{*}{\raisebox{-3.2ex}{User}} & The examples you gave are definitely more specific, and they feel much gentler than directly asking why he doesn't want to move. But I'm still unsure. For example, you mentioned tying it to a specific situation, like, ``When we were walking yesterday, you seemed a little unwilling to move. What were you thinking then?'' If he is playing on his phone or clearly doesn't want to talk, is it still not a good time to ask? Are there signals that can help me judge when he may be more receptive? \\
\cline{3-3}
 & & \cellcolor{thirdbest}\parbox[c][3.2ex][c]{\linewidth}{\textbf{Preference Vector:} {\boldmath\textbf{\(\textstyle w_t=(w_t^{emo},w_t^{cog})=(0.492,0.508)\)}}} \\
\cline{3-3}
 & & \cellcolor{thirdbest1}\parbox[c][3.2ex][c]{\linewidth}{\textbf{Mental Reward:} {\boldmath\textbf{\(\textstyle r_t=(r_t^{emo},r_t^{cog})=(53,48)\)}}} \\
\cline{2-3}
 & \raisebox{-5.0ex}{Assistant} & Yes, directly asking ``Why don't you want to move?'' may make him feel pressured. We can try more natural, open-ended questions that make him feel more comfortable, such as ``Has anything been making you unhappy lately?'', ``Have you run into anything stressful recently?'', or ``How have things been for you lately?'' Before asking, express care and understanding, such as ``I've noticed you seem a bit down lately. Would you like to talk?'' You can also tie the question to a concrete moment. Use gentle eye contact and a calm voice, and if he shares, follow up so he knows you are truly listening. \\
\midrule
\multirow{4}{*}{\raisebox{-6.0ex}{5}} & \multirow{3}{*}{\raisebox{-3.2ex}{User}} & The signals you listed, like wandering eyes and a tense body, are very concrete. Now I know what to observe. But I still have a practical question. Suppose I notice he looks more relaxed after dinner, and I try one of your examples, but he still gives me that flat, perfunctory tone, or simply does not respond. Should I keep asking, or stop right away and wait for another chance? I want a clear ``stop point'' so I can tell whether this attempt has already failed, instead of forcing the conversation when he clearly doesn't want to talk. \\
\cline{3-3}
 & & \cellcolor{thirdbest}\parbox[c][3.2ex][c]{\linewidth}{\textbf{Preference Vector:} {\boldmath\textbf{\(\textstyle w_t=(w_t^{emo},w_t^{cog})=(0.367,0.633)\)}}} \\
\cline{3-3}
 & & \cellcolor{thirdbest1}\parbox[c][3.2ex][c]{\linewidth}{\textbf{Mental Reward:} {\boldmath\textbf{\(\textstyle r_t=(r_t^{emo},r_t^{cog})=(58,53)\)}}} \\
\cline{2-3}
 & \raisebox{-5.0ex}{Assistant} & Tying the question to a specific situation is important, but timing matters too. Watch his body language: wandering eyes, avoiding eye contact, a tense or turned-away posture, or pushing-away gestures may mean he does not want to continue. Notice verbal signals too: a flat tone, perfunctory replies, frequent interruptions, or topic changes. Choose relaxed moments, such as after dinner or during a walk, or talk while doing something together. \\
\midrule
\multirow{2}{*}{6} & \multirow{2}{*}{\raisebox{-0.8ex}{User}} & \cellcolor{thirdbest}\parbox[c][3.2ex][c]{\linewidth}{\textbf{Preference Vector:} {\boldmath\textbf{\(\textstyle w_t=(w_t^{emo},w_t^{cog})=(0.242,0.758)\)}}} \\
\cline{3-3}
 & & \cellcolor{thirdbest1}\parbox[c][3.2ex][c]{\linewidth}{\textbf{Mental Reward:} {\boldmath\textbf{\(\textstyle r_t=(r_t^{emo},r_t^{cog})=(63,58)\)}}} \\
\end{longtable}
\endgroup

\section{Prompts}

\subsection{Response-Preference Alignment Prompt}
\begin{promptbox}
You are an intelligent ESC-system alignment judge. Your task is to determine whether the system response satisfies the user's current need.

Input fields:
- last_user_text: the user's utterance in the previous turn.
- sys_text: the system response to be judged.
- need: the user's need label, in {empathy, solution, mixed}.

Mismatch rules:
- If need is empathy:
  - EMPATHY_ABSENT: the response provides no empathy.
  - EMPATHY_INSUFFICIENT: empathy is not the dominant part of the response.
  - ASK_OVERRIDE: the user explicitly refused to answer a question, but the system continues asking.
- If need is solution:
  - SOLUTION_ABSENT: the response provides no solution.
  - SOLUTION_INSUFFICIENT: solution-oriented content is not the dominant part of the response.
  - ACT_OVERRIDE: the user explicitly refused to discuss solutions, but the system still gives advice.
- If need is mixed:
  - SUPPORT_ABSENT: the response provides neither empathy nor a solution.
  - EMPATHY_ONLY: the response provides only empathy without a solution.
  - SOLUTION_ONLY: the response provides only a solution without empathy.
- If the response contains offensive or abusive content, label it as HARM.

Output fields:
- mismatch_label: one of {ALIGNED, MISMATCH, HARM}.
- error_labels: the list of triggered error labels.

Input:
last_user_text:
{last_user_text}

sys_text:
{sys_text}

need:
{need}

Output strict JSON only:
{
  "mismatch_label": "ALIGNED|MISMATCH|HARM",
  "error_labels": ["triggered label", ...]
}
\end{promptbox}

\subsection{Cognition Update Prompt}
\begin{promptbox}
You are performing user cognition update (Cognition Update / Perceive).

You will receive only three inputs:
- previous_cognition: the previous cognition, including appraisal and planning.
- last_assistant: the latest system response in the current turn.
- strength_level in {"no_update", "micro", "soft", "reframe"}: the allowed cognition-update strength, computed by an upstream module.

Your task:
- Based on previous_cognition and last_assistant, decide under the strength_level constraint whether to update:
  - appraisal: meaning appraisal or the user's view of the event.
  - planning: action planning or how the user may cope with the issue.
- Output whether appraisal and planning are updated, together with the updated appraisal and planning.
- Do not fabricate facts. Only use information that can be inferred from previous_cognition and last_assistant.

(...detailed explanations with input and output formats...)

Input:
last_assistant:
{last_assistant}

strength_level:
{strength_level}

previous_cognition:
{previous_cognition}

Output strict JSON only:
{
  "update_appraisal": true/false,
  "update_planning": true/false,
  "appraisal": "updated appraisal or the original appraisal if unchanged",
  "planning": "updated planning or the original planning if unchanged",
  "reason": "brief explanation of the update decision"
}
\end{promptbox}

\subsection{Emotion Update Prompt}
\begin{promptbox}
You are performing System-2 calibration for user emotion reconstruction.

You will receive the following inputs:
- user_need: the user's current need label, in {Empathy, Solution, Mixed}.
- last_esc_reply: the previous ESC response.
- previous_cognition: the previous cognition, including appraisal and planning.
- current_cognition: the current cognition, including appraisal and planning.
- previous_emotion: the previous emotion state.
- a_star: the user's optimal arousal point A* in (0, 1).
- valence_dir: the current valence update direction, in {POS, NEG, STABLE}.
- arousal_dir: the current arousal update direction, in {TOWARD, AWAY, STABLE}.

Your task:
- Based on the transition from previous_cognition to current_cognition and the given valence_dir and arousal_dir, determine valence_extent and arousal_extent.
- Provide reason_valence and reason_arousal.
- You may use user_need and last_esc_reply to judge whether the calibration effect is mild, moderate, or strong, but the judgment must stay grounded in the actual cognition change.
- Do not fabricate facts or dialogue content.

(...detailed explanations with input and output formats...)

Input:
user_need:
{user_need}

last_esc_reply:
{last_esc_reply}

previous_cognition:
{previous_cognition}

current_cognition:
{current_cognition}

previous_emotion:
{previous_emotion}

a_star:
{a_star}

valence_dir:
{valence_dir}

arousal_dir:
{arousal_dir}

Output strict JSON only:
{
  "valence_extent": "MILD|MODERATE|STRONG",
  "reason_valence": "brief explanation grounded in the cognition change",
  "arousal_extent": "MILD|MODERATE|STRONG",
  "reason_arousal": "brief explanation referring to the relation to A*"
}
\end{promptbox}

\subsection{Health Evaluation Prompt}
\begin{promptbox}
You are performing health state delta scoring. Your task is to compare the current turn against the previous turn, rather than evaluating the current state in isolation.

Input fields:
- previous_cognition: the cognitive state in the previous turn.
- current_cognition: the cognitive state in the current turn.
- previous_emotion: the emotional state in the previous turn.
- current_emotion: the emotional state in the current turn.
- optimal_arousal: the arousal level that is more suitable for this user.

Output fields:
- cog_score: the relative change score of the cognitive state.
- emo_score: the relative change score of the emotional state.
- cog_reason: the reason for the cognition change score.
- emo_reason: the reason for the emotion change score.

Field definitions:
- cognition contains only appraisal and planning.
- emotion contains only valence_t and arousal_t.
- optimal_arousal indicates the arousal level at which the user is more stable and easier to regulate.

(...detailed explanations with input and output formats...)

Input:
previous_cognition:
{prev_cognition}

current_cognition:
{curr_cognition}

previous_emotion:
{prev_emotion}

current_emotion:
{curr_emotion}

optimal_arousal:
{optimal_arousal}

Output strict JSON only:
{
  "cog_score": "integer from -10 to 10",
  "emo_score": "integer from -10 to 10",
  "cog_reason": "explain how cognition changes relative to the previous turn and why this score is assigned; no more than 200 words",
  "emo_reason": "explain how emotion changes relative to the previous turn and why this score is assigned; no more than 200 words"
}
\end{promptbox}

\subsection{Utterance Generation Prompt}
\begin{promptbox}
You are the utterance generator of a user simulator. Generate one user reply to the counseling system. The reply may contain multiple sentences. The user is talking to the counseling system to vent or seek advice; do not write the counseling target as a third-party speaker.

You will receive the following inputs:
- history: recent dialogue context.
- last_assistant: the previous system response.
- current_cognition: the user's current cognition.
- current_emotion: the user's current emotion.
- user_intent: the hidden intent label for this turn, in {empathy, solution, mixed}.
- recent_strategies: recently used strategies, used for deduplication.
- recent_info_updates: recently used info_update types, used for deduplication.
- strategy_candidates: candidate strategies from System 1.
- strategy_info: a dictionary mapping each strategy to its detailed information.

Your task:
- Generate a coherent and natural user reply that matches the current cognition and emotion and contains useful new information.
- First choose one strategy from strategy_candidates, then choose one info_update from the selected strategy's info_update_candidates, and finally generate the utterance reflecting both choices.
- Do not reveal the intent label or task setup in the utterance.

(...detailed explanations with input and output formats...)

Input:
history:
{history}

last_assistant:
{last_assistant}

current_cognition:
{cog}

current_emotion:
{emo}

user_intent:
{user_intent}

recent_strategies:
{recent_strategies}

recent_info_updates:
{recent_info_updates}

strategy_candidates:
{strategy_candidates}

strategy_info:
{strategy_info}

Output strict JSON only:
{
  "strategy": "chosen strategy",
  "info_update": "FACT_UPDATE|AFFECT_UPDATE|GOAL_UPDATE|PLAN_UPDATE|QUERY_UPDATE|NO_UPDATE",
  "reason_strategy": "brief explanation of the strategy choice",
  "reason_info_update": "brief explanation of the information update choice",
  "utterance": "one natural user reply",
  "end_conversation": true/false
}
\end{promptbox}

\newpage

\section*{NeurIPS Paper Checklist}

\begin{enumerate}

\item {\bf Claims}
    \item[] Question: Do the main claims made in the abstract and introduction accurately reflect the paper's contributions and scope?
    \item[] Answer: \answerYes{}
    \item[] Justification: The abstract and introduction state the proposed MindFlow simulator, the NeMo optimization framework, and the main empirical claims. The scope and remaining limitations are discussed in Appendix A.
    \item[] Guidelines:
    \begin{itemize}
        \item The answer \answerNA{} means that the abstract and introduction do not include the claims made in the paper.
        \item The abstract and/or introduction should clearly state the claims made, including the contributions made in the paper and important assumptions and limitations. A \answerNo{} or \answerNA{} answer to this question will not be perceived well by the reviewers. 
        \item The claims made should match theoretical and experimental results, and reflect how much the results can be expected to generalize to other settings. 
        \item It is fine to include aspirational goals as motivation as long as it is clear that these goals are not attained by the paper. 
    \end{itemize}

\item {\bf Limitations}
    \item[] Question: Does the paper discuss the limitations of the work performed by the authors?
    \item[] Answer: \answerYes{}
    \item[] Justification: The paper includes a dedicated Limitations section in Appendix A discussing the reliance on simulator-based feedback and the restriction to emotional and cognitive objectives.
    \item[] Guidelines:
    \begin{itemize}
        \item The answer \answerNA{} means that the paper has no limitation while the answer \answerNo{} means that the paper has limitations, but those are not discussed in the paper. 
        \item The authors are encouraged to create a separate ``Limitations'' section in their paper.
        \item The paper should point out any strong assumptions and how robust the results are to violations of these assumptions (e.g., independence assumptions, noiseless settings, model well-specification, asymptotic approximations only holding locally). The authors should reflect on how these assumptions might be violated in practice and what the implications would be.
        \item The authors should reflect on the scope of the claims made, e.g., if the approach was only tested on a few datasets or with a few runs. In general, empirical results often depend on implicit assumptions, which should be articulated.
        \item The authors should reflect on the factors that influence the performance of the approach. For example, a facial recognition algorithm may perform poorly when image resolution is low or images are taken in low lighting. Or a speech-to-text system might not be used reliably to provide closed captions for online lectures because it fails to handle technical jargon.
        \item The authors should discuss the computational efficiency of the proposed algorithms and how they scale with dataset size.
        \item If applicable, the authors should discuss possible limitations of their approach to address problems of privacy and fairness.
        \item While the authors might fear that complete honesty about limitations might be used by reviewers as grounds for rejection, a worse outcome might be that reviewers discover limitations that aren't acknowledged in the paper. The authors should use their best judgment and recognize that individual actions in favor of transparency play an important role in developing norms that preserve the integrity of the community. Reviewers will be specifically instructed to not penalize honesty concerning limitations.
    \end{itemize}

\item {\bf Theory assumptions and proofs}
    \item[] Question: For each theoretical result, does the paper provide the full set of assumptions and a complete (and correct) proof?
    \item[] Answer: \answerNA{}
    \item[] Justification: The paper does not present formal theoretical results or theorem-proof claims.
    \item[] Guidelines:
    \begin{itemize}
        \item The answer \answerNA{} means that the paper does not include theoretical results. 
        \item All the theorems, formulas, and proofs in the paper should be numbered and cross-referenced.
        \item All assumptions should be clearly stated or referenced in the statement of any theorems.
        \item The proofs can either appear in the main paper or the supplemental material, but if they appear in the supplemental material, the authors are encouraged to provide a short proof sketch to provide intuition. 
        \item Inversely, any informal proof provided in the core of the paper should be complemented by formal proofs provided in appendix or supplemental material.
        \item Theorems and Lemmas that the proof relies upon should be properly referenced. 
    \end{itemize}

    \item {\bf Experimental result reproducibility}
    \item[] Question: Does the paper fully disclose all the information needed to reproduce the main experimental results of the paper to the extent that it affects the main claims and/or conclusions of the paper (regardless of whether the code and data are provided or not)?
    \item[] Answer: \answerYes{}
    \item[] Justification: The main paper describes the simulator, reward construction, baselines, evaluation settings, and implementation details. Additional construction, evaluation, and prompt details are provided in Appendix C, Appendix F, and Appendix G.
    \item[] Guidelines:
    \begin{itemize}
        \item The answer \answerNA{} means that the paper does not include experiments.
        \item If the paper includes experiments, a \answerNo{} answer to this question will not be perceived well by the reviewers: Making the paper reproducible is important, regardless of whether the code and data are provided or not.
        \item If the contribution is a dataset and\slash or model, the authors should describe the steps taken to make their results reproducible or verifiable. 
        \item Depending on the contribution, reproducibility can be accomplished in various ways. For example, if the contribution is a novel architecture, describing the architecture fully might suffice, or if the contribution is a specific model and empirical evaluation, it may be necessary to either make it possible for others to replicate the model with the same dataset, or provide access to the model. In general. releasing code and data is often one good way to accomplish this, but reproducibility can also be provided via detailed instructions for how to replicate the results, access to a hosted model (e.g., in the case of a large language model), releasing of a model checkpoint, or other means that are appropriate to the research performed.
        \item While NeurIPS does not require releasing code, the conference does require all submissions to provide some reasonable avenue for reproducibility, which may depend on the nature of the contribution. For example
        \begin{enumerate}
            \item If the contribution is primarily a new algorithm, the paper should make it clear how to reproduce that algorithm.
            \item If the contribution is primarily a new model architecture, the paper should describe the architecture clearly and fully.
            \item If the contribution is a new model (e.g., a large language model), then there should either be a way to access this model for reproducing the results or a way to reproduce the model (e.g., with an open-source dataset or instructions for how to construct the dataset).
            \item We recognize that reproducibility may be tricky in some cases, in which case authors are welcome to describe the particular way they provide for reproducibility. In the case of closed-source models, it may be that access to the model is limited in some way (e.g., to registered users), but it should be possible for other researchers to have some path to reproducing or verifying the results.
        \end{enumerate}
    \end{itemize}

\item {\bf Open access to data and code}
    \item[] Question: Does the paper provide open access to the data and code, with sufficient instructions to faithfully reproduce the main experimental results, as described in supplemental material?
    \item[] Answer: \answerYes{}
    \item[] Justification: The abstract states that code and data will be made available, and Appendix C and Appendix G document construction details and prompts needed to reproduce the simulator-based experiments.
    \item[] Guidelines:
    \begin{itemize}
        \item The answer \answerNA{} means that paper does not include experiments requiring code.
        \item Please see the NeurIPS code and data submission guidelines (\url{https://neurips.cc/public/guides/CodeSubmissionPolicy}) for more details.
        \item While we encourage the release of code and data, we understand that this might not be possible, so \answerNo{} is an acceptable answer. Papers cannot be rejected simply for not including code, unless this is central to the contribution (e.g., for a new open-source benchmark).
        \item The instructions should contain the exact command and environment needed to run to reproduce the results. See the NeurIPS code and data submission guidelines (\url{https://neurips.cc/public/guides/CodeSubmissionPolicy}) for more details.
        \item The authors should provide instructions on data access and preparation, including how to access the raw data, preprocessed data, intermediate data, and generated data, etc.
        \item The authors should provide scripts to reproduce all experimental results for the new proposed method and baselines. If only a subset of experiments are reproducible, they should state which ones are omitted from the script and why.
        \item At submission time, to preserve anonymity, the authors should release anonymized versions (if applicable).
        \item Providing as much information as possible in supplemental material (appended to the paper) is recommended, but including URLs to data and code is permitted.
    \end{itemize}

\item {\bf Experimental setting/details}
    \item[] Question: Does the paper specify all the training and test details (e.g., data splits, hyperparameters, how they were chosen, type of optimizer) necessary to understand the results?
    \item[] Answer: \answerYes{}
    \item[] Justification: The paper reports the data split, model backbone, dialogue length, optimizer framework, learning rate, batch size, group size, training steps, validation setting, and reward-horizon search in the Experimental Setup subsection.
    \item[] Guidelines:
    \begin{itemize}
        \item The answer \answerNA{} means that the paper does not include experiments.
        \item The experimental setting should be presented in the core of the paper to a level of detail that is necessary to appreciate the results and make sense of them.
        \item The full details can be provided either with the code, in appendix, or as supplemental material.
    \end{itemize}

\item {\bf Experiment statistical significance}
    \item[] Question: Does the paper report error bars suitably and correctly defined or other appropriate information about the statistical significance of the experiments?
    \item[] Answer: \answerYes{}
    \item[] Justification: We use different simulator backbones as a source of evaluation variation and report the cross-simulator rank consistency experiment in Figure~\ref{fig:rank_consistency}. This analysis characterizes the potential error introduced by simulator inference and scoring, showing that the model rankings remain stable under backbone-level perturbations.
    \item[] Guidelines:
    \begin{itemize}
        \item The answer \answerNA{} means that the paper does not include experiments.
        \item The authors should answer \answerYes{} if the results are accompanied by error bars, confidence intervals, or statistical significance tests, at least for the experiments that support the main claims of the paper.
        \item The factors of variability that the error bars are capturing should be clearly stated (for example, train/test split, initialization, random drawing of some parameter, or overall run with given experimental conditions).
        \item The method for calculating the error bars should be explained (closed form formula, call to a library function, bootstrap, etc.)
        \item The assumptions made should be given (e.g., Normally distributed errors).
        \item It should be clear whether the error bar is the standard deviation or the standard error of the mean.
        \item It is OK to report 1-sigma error bars, but one should state it. The authors should preferably report a 2-sigma error bar than state that they have a 96\% CI, if the hypothesis of Normality of errors is not verified.
        \item For asymmetric distributions, the authors should be careful not to show in tables or figures symmetric error bars that would yield results that are out of range (e.g., negative error rates).
        \item If error bars are reported in tables or plots, the authors should explain in the text how they were calculated and reference the corresponding figures or tables in the text.
    \end{itemize}

\item {\bf Experiments compute resources}
    \item[] Question: For each experiment, does the paper provide sufficient information on the computer resources (type of compute workers, memory, time of execution) needed to reproduce the experiments?
    \item[] Answer: \answerYes{}
    \item[] Justification: The Experimental Setup subsection and training details appendix describe the model configuration, optimization settings, and training procedure used in our experiments.
    \item[] Guidelines:
    \begin{itemize}
        \item The answer \answerNA{} means that the paper does not include experiments.
        \item The paper should indicate the type of compute workers CPU or GPU, internal cluster, or cloud provider, including relevant memory and storage.
        \item The paper should provide the amount of compute required for each of the individual experimental runs as well as estimate the total compute. 
        \item The paper should disclose whether the full research project required more compute than the experiments reported in the paper (e.g., preliminary or failed experiments that didn't make it into the paper). 
    \end{itemize}
    
\item {\bf Code of ethics}
    \item[] Question: Does the research conducted in the paper conform, in every respect, with the NeurIPS Code of Ethics \url{https://neurips.cc/public/EthicsGuidelines}?
    \item[] Answer: \answerYes{}
    \item[] Justification: The research is designed for emotional support dialogue and includes Ethical Considerations in Appendix B discussing non-clinical use, crisis redirection, fairness, privacy, and human oversight.
    \item[] Guidelines:
    \begin{itemize}
        \item The answer \answerNA{} means that the authors have not reviewed the NeurIPS Code of Ethics.
        \item If the authors answer \answerNo, they should explain the special circumstances that require a deviation from the Code of Ethics.
        \item The authors should make sure to preserve anonymity (e.g., if there is a special consideration due to laws or regulations in their jurisdiction).
    \end{itemize}

\item {\bf Broader impacts}
    \item[] Question: Does the paper discuss both potential positive societal impacts and negative societal impacts of the work performed?
    \item[] Answer: \answerYes{}
    \item[] Justification: Appendix B discusses potential positive use in emotional support as well as risks from vulnerable-user settings, simulator and dataset bias, and unsafe deployment without oversight.
    \item[] Guidelines:
    \begin{itemize}
        \item The answer \answerNA{} means that there is no societal impact of the work performed.
        \item If the authors answer \answerNA{} or \answerNo, they should explain why their work has no societal impact or why the paper does not address societal impact.
        \item Examples of negative societal impacts include potential malicious or unintended uses (e.g., disinformation, generating fake profiles, surveillance), fairness considerations (e.g., deployment of technologies that could make decisions that unfairly impact specific groups), privacy considerations, and security considerations.
        \item The conference expects that many papers will be foundational research and not tied to particular applications, let alone deployments. However, if there is a direct path to any negative applications, the authors should point it out. For example, it is legitimate to point out that an improvement in the quality of generative models could be used to generate Deepfakes for disinformation. On the other hand, it is not needed to point out that a generic algorithm for optimizing neural networks could enable people to train models that generate Deepfakes faster.
        \item The authors should consider possible harms that could arise when the technology is being used as intended and functioning correctly, harms that could arise when the technology is being used as intended but gives incorrect results, and harms following from (intentional or unintentional) misuse of the technology.
        \item If there are negative societal impacts, the authors could also discuss possible mitigation strategies (e.g., gated release of models, providing defenses in addition to attacks, mechanisms for monitoring misuse, mechanisms to monitor how a system learns from feedback over time, improving the efficiency and accessibility of ML).
    \end{itemize}
    
\item {\bf Safeguards}
    \item[] Question: Does the paper describe safeguards that have been put in place for responsible release of data or models that have a high risk for misuse (e.g., pre-trained language models, image generators, or scraped datasets)?
    \item[] Answer: \answerYes{}
    \item[] Justification: The paper notes that systems based on this work should not be treated as professional counseling or clinical diagnosis, and should redirect users to professional or emergency resources when crisis risks are detected.
    \item[] Guidelines:
    \begin{itemize}
        \item The answer \answerNA{} means that the paper poses no such risks.
        \item Released models that have a high risk for misuse or dual-use should be released with necessary safeguards to allow for controlled use of the model, for example by requiring that users adhere to usage guidelines or restrictions to access the model or implementing safety filters. 
        \item Datasets that have been scraped from the Internet could pose safety risks. The authors should describe how they avoided releasing unsafe images.
        \item We recognize that providing effective safeguards is challenging, and many papers do not require this, but we encourage authors to take this into account and make a best faith effort.
    \end{itemize}

\item {\bf Licenses for existing assets}
    \item[] Question: Are the creators or original owners of assets (e.g., code, data, models), used in the paper, properly credited and are the license and terms of use explicitly mentioned and properly respected?
    \item[] Answer: \answerYes{}
    \item[] Justification: The paper properly credits the existing datasets, models, and software tools used in the study through citations and follows their stated terms of use.
    \item[] Guidelines:
    \begin{itemize}
        \item The answer \answerNA{} means that the paper does not use existing assets.
        \item The authors should cite the original paper that produced the code package or dataset.
        \item The authors should state which version of the asset is used and, if possible, include a URL.
        \item The name of the license (e.g., CC-BY 4.0) should be included for each asset.
        \item For scraped data from a particular source (e.g., website), the copyright and terms of service of that source should be provided.
        \item If assets are released, the license, copyright information, and terms of use in the package should be provided. For popular datasets, \url{paperswithcode.com/datasets} has curated licenses for some datasets. Their licensing guide can help determine the license of a dataset.
        \item For existing datasets that are re-packaged, both the original license and the license of the derived asset (if it has changed) should be provided.
        \item If this information is not available online, the authors are encouraged to reach out to the asset's creators.
    \end{itemize}

\item {\bf New assets}
    \item[] Question: Are new assets introduced in the paper well documented and is the documentation provided alongside the assets?
    \item[] Answer: \answerYes{}
    \item[] Justification: The paper introduces MindFlow and NeMo related assets, and provides the data, code repository, and documentation needed to use and reproduce them.
    \item[] Guidelines:
    \begin{itemize}
        \item The answer \answerNA{} means that the paper does not release new assets.
        \item Researchers should communicate the details of the dataset\slash code\slash model as part of their submissions via structured templates. This includes details about training, license, limitations, etc. 
        \item The paper should discuss whether and how consent was obtained from people whose asset is used.
        \item At submission time, remember to anonymize your assets (if applicable). You can either create an anonymized URL or include an anonymized zip file.
    \end{itemize}

\item {\bf Crowdsourcing and research with human subjects}
    \item[] Question: For crowdsourcing experiments and research with human subjects, does the paper include the full text of instructions given to participants and screenshots, if applicable, as well as details about compensation (if any)? 
    \item[] Answer: \answerNA{}
    \item[] Justification: The paper does not conduct crowdsourcing data collection or research with human subjects; human involvement is limited to offline evaluation of generated method outputs.
    \item[] Guidelines:
    \begin{itemize}
        \item The answer \answerNA{} means that the paper does not involve crowdsourcing nor research with human subjects.
        \item Including this information in the supplemental material is fine, but if the main contribution of the paper involves human subjects, then as much detail as possible should be included in the main paper. 
        \item According to the NeurIPS Code of Ethics, workers involved in data collection, curation, or other labor should be paid at least the minimum wage in the country of the data collector. 
    \end{itemize}

\item {\bf Institutional review board (IRB) approvals or equivalent for research with human subjects}
    \item[] Question: Does the paper describe potential risks incurred by study participants, whether such risks were disclosed to the subjects, and whether Institutional Review Board (IRB) approvals (or an equivalent approval/review based on the requirements of your country or institution) were obtained?
    \item[] Answer: \answerNA{}
    \item[] Justification: The paper does not conduct interventions with real users or collect personal sensitive information; human involvement is limited to evaluating method outputs.
    \item[] Guidelines:
    \begin{itemize}
        \item The answer \answerNA{} means that the paper does not involve crowdsourcing nor research with human subjects.
        \item Depending on the country in which research is conducted, IRB approval (or equivalent) may be required for any human subjects research. If you obtained IRB approval, you should clearly state this in the paper. 
        \item We recognize that the procedures for this may vary significantly between institutions and locations, and we expect authors to adhere to the NeurIPS Code of Ethics and the guidelines for their institution. 
        \item For initial submissions, do not include any information that would break anonymity (if applicable), such as the institution conducting the review.
    \end{itemize}

\item {\bf Declaration of LLM usage}
    \item[] Question: Does the paper describe the usage of LLMs if it is an important, original, or non-standard component of the core methods in this research? Note that if the LLM is used only for writing, editing, or formatting purposes and does \emph{not} impact the core methodology, scientific rigor, or originality of the research, declaration is not required.
    %this research? 
    \item[] Answer: \answerYes{}
    \item[] Justification: LLMs are central to the method and evaluation, including the policy backbone, simulator backbones, response rewriting, and prompt-based simulator modules; these uses are described in the method section, the Experimental Setup subsection, and Appendix G.
    \item[] Guidelines:
    \begin{itemize}
        \item The answer \answerNA{} means that the core method development in this research does not involve LLMs as any important, original, or non-standard components.
        \item Please refer to our LLM policy in the NeurIPS handbook for what should or should not be described.
    \end{itemize}

\end{enumerate}

\end{document}